%% file: icml_main.tex
\documentclass{article}

\usepackage{microtype}
\usepackage{graphicx}
\usepackage{subfigure}
\usepackage{booktabs} %

\usepackage{hyperref}

\usepackage[accepted]{icml2024}
\usepackage{algorithm}

\usepackage{amsmath}
\usepackage{amssymb}
\usepackage{mathtools}
\usepackage{amsthm}

\usepackage[capitalize,noabbrev]{cleveref}

\theoremstyle{plain}

\theoremstyle{definition}

\theoremstyle{remark}

\usepackage[textsize=tiny]{todonotes}

\def\titlename{GenCO: \textbf{Gen}erating Diverse Designs with \textbf{Co}mbinatorial Constraints}

\icmltitlerunning{\titlename}

\usepackage{amsfonts}       %
\usepackage{nicefrac}       %
\usepackage{amsmath,amssymb}
\usepackage{graphicx}
\usepackage{booktabs}
\usepackage{multirow}
\usepackage{pifont}

\graphicspath{{grf/}, {grf/warcraft}}

\input{00_math_commands.tex}

\input{00_definitions.tex}

\begin{document}

\twocolumn[
\icmltitle{\titlename}

\icmlsetaffilorder{usc,cornell,FAIR}

\begin{icmlauthorlist}
\icmlauthor{Aaron M. Ferber}{usc,cornell} 
\icmlauthor{Arman Zharmagambetov}{FAIR}
\icmlauthor{Taoan Huang}{usc}
\icmlauthor{Bistra Dilkina}{usc}
\icmlauthor{Yuandong Tian}{FAIR}
\end{icmlauthorlist}

\icmlaffiliation{usc}{Department of Computer Science, University of Southern California, Los Angeles, CA, USA}
\icmlaffiliation{cornell}{Department of Computer Science, Cornell University, Ithaca, NY, USA}
\icmlaffiliation{FAIR}{AI at Meta (FAIR), Menlo Park, CA, USA}

\icmlcorrespondingauthor{Aaron M. Ferber}{aferber@usc.edu}
\icmlcorrespondingauthor{Yuandong Tian}{yuandong@meta.com}

\icmlkeywords{Machine Learning, ICML}

\vskip 0.3in
]

\printAffiliationsAndNotice{\icmlEqualContribution} %

\begin{abstract}
Deep generative models like GAN and VAE have shown impressive results in generating unconstrained objects like images. 
However, many design settings arising in industrial design, material science, computer graphics and more require that the generated objects satisfy hard combinatorial constraints or meet objectives in addition to modeling a data distribution. 
To address this, we propose GenCO, a generative framework that guarantees constraint satisfaction throughout training by leveraging differentiable combinatorial solvers to enforce feasibility.
GenCO imposes the generative loss on provably feasible solutions rather than intermediate soft solutions, meaning that the deep generative network can focus on ensuring the generated objects match the data distribution without having to also capture feasibility.
This shift enables practitioners to enforce hard constraints on the generated outputs during end-to-end training, enabling assessments of their feasibility and introducing additional combinatorial loss components to deep generative training.
We demonstrate the effectiveness of our approach on a variety of generative combinatorial tasks, including game level generation, map creation for path planning, and photonic device design, consistently demonstrating its capability to yield diverse, high-quality solutions that verifiably adhere to user-specified combinatorial properties.

\end{abstract}

\input{content}

\bibliography{references}
\bibliographystyle{icml2024}

\newpage
\appendix
\onecolumn

\input{appendix}

\end{document}

%% file: 00_math_commands.tex
\usepackage{amsmath,amsfonts,bm}

\def\eqref#1{equation~\ref{#1}}

\def\1{\bm{1}}
\newcommand{\train}{\mathcal{D}}

\def\vc{{\bm{c}}}

\def\vg{{\bm{g}}}

\def\vp{{\bm{p}}}

\def\vx{{\bm{x}}}
\def\vy{{\bm{y}}}

\DeclareMathAlphabet{\mathsfit}{\encodingdefault}{\sfdefault}{m}{sl}
\SetMathAlphabet{\mathsfit}{bold}{\encodingdefault}{\sfdefault}{bx}{n}

\newcommand{\E}{\mathbb{E}}

\DeclareMathOperator*{\argmax}{arg\,max}
\DeclareMathOperator*{\argmin}{arg\,min}

%% file: 00_definitions.tex
\newboolean{showcomments}
\setboolean{showcomments}{true} %
\ifthenelse{\boolean{showcomments}}
  {\newcommand{\nb}[2]{
    \fcolorbox{gray}{yellow}{\bfseries\sffamily\scriptsize#1}
    {$\blacktriangleright$#2$\blacktriangleleft$}
   }
   
  }
  {\newcommand{\nb}[2]{}
   
  }

\newcommand \yuandong[1]{\nb{YD}{\textcolor{brown}{\textsl{#1}}}} 

\newcommand{\genco}{GenCO} %

\newcommand{\feas}{\Omega} %
\newcommand{\noise}{\epsilon} %
\newcommand{\genloss}{\mathcal{L}} %
\newcommand{\combloss}{\mathcal{C}} %
\newcommand{\gengenparams}{\theta} %

\newcommand{\unconstrainedgen}{\tilde{G}} %
\newcommand{\solver}{S} %

%% file: content.tex
\newcommand{\cmark}{\ding{51}}
\newcommand{\xmark}{\ding{55}}

\section{Introduction}
Generating diverse and realistic objects with combinatorial properties is an important task with many applications. For AI-guided design in particular, the goal is to generate a diverse set of combinatorial solutions that are both feasible and high quality with respect to a given nonlinear objective. Diversity is often useful when dealing with uncertain metrics, diverse preferences, or for the sake of further manual scrutiny. For example, in photonic device design, we want to generate a variety of devices that meet foundry manufacturing constraints and optimize physics-related objectives \citep{schubert2022inverse}. 
In the design of new molecules, we want to generate a variety of chemically valid molecules with specific properties \citep{pereira2021moleculediversity}. In other scenarios, we would like to create a diverse set of designs so that their solutions (not necessarily diverse given the design) meet certain objectives. For example, in video game level design, we may want to generate levels that are both realistic and valid/playable \citep{zhang2020milpgan}. Here, ``valid'' may refer to discrete characteristics of the level that must be satisfied (e.g., a minimum number of enemies, a guaranteed path between the level entrance and exit, etc.).

\def\cL{\mathcal{L}}
\def\cC{\mathcal{C}}
\def\cD{\mathcal{D}}
\def\cX{\mathcal{X}}

\begin{table*}
\small
    \centering
    \begin{tabular}{r|cccc}
      \toprule
                  \multirow{2}{*}{Approaches}  
                  & Solution
                  & Feasibility
                  & Data
                  & Nonlinear \\  
                  &  generation & guarantees & distribution & objective \\
     \midrule
     GAN/VAE~\cite{goodfellow2014gan,kingma2013vae} &\textbf{Fast} & \xmark & \cmark & \cmark \\
     Integer Programming~\cite{deaton2014survey} & Slow & \cmark & \xmark & \xmark \\
     Lagrangian Multiplier~\cite{di2020can}    &  Slow & \xmark & \cmark & \cmark \\  
     SurCo~\cite{ferber2023surco} & Slow & \cmark & \xmark & \cmark \\
     Fixing/Rejection~\cite{zhang2020milpgan,gomes2006rejection} & Slow & \cmark & \xmark & \cmark \\
     \midrule
     
     \textbf{\genco{}} (proposed) & \textbf{Fast} & \cmark & \cmark & \cmark \\
     
     \bottomrule
    \end{tabular}
    \caption{A comparison of relevant directions tackling generative modeling or combinatorial optimization for design problems.}
    \label{tab:summary_comparison}
\end{table*}

\def\gen{\mathrm{gen}}
\def\ind{\mathrm{ind}}

Mathematically, we seek several solutions $\cX := \{\vx_j\}$ to the following optimization problem:
\begin{equation}
   \min_{\cX} \cL(\cX) + \gamma \sum_j \cD(\vx_j) \quad\quad \mathrm{s.t.}\ \vx_j\in\Omega \label{eq:formulation} 
\end{equation}
where $\cL$ is a \emph{group loss} defined on a population of solutions $\cX$ (e.g., minimizing the Wasserstein distance to a dataset), $\cD$ is an \emph{individual loss} measuring the quality of each solution $\vx_j$, $\gamma$ is the weighing coefficient and $\Omega$ is the feasible set. Eqn.~\ref{eq:formulation} can be solved by direct optimization or learning approaches that leverage solution datasets.  
 
Several approaches can attempt to solve this problem. 
The \emph{Lagrangian}~\cite{frangioni2005lagrangian} can reformulate the problem as $\min_{\cX} \cL(\cX) + \gamma\sum_j \cD(\vx_j) + \lambda \sum_j \mathrm{dist}_\Omega(\vx_j)$, where $\mathrm{dist}_\Omega(\vx_j)$ measures the distance between $\vx_j$ and the feasible set $\Omega$ ($\mathrm{dist}_\Omega(\vx_j) = 0$ if $\vx_j\in \Omega$). However the final solutions $\vx_j$ may not be feasible unless we use $\lambda \rightarrow +\infty$ which may lead to numerical instability. 
To reduce the optimization cost, we want to learn a \emph{generative model} to directly generate $\vx_j$, and avoid on-the-fly optimization. However, standard methods like GAN~\citep{goodfellow2014gan} or VAE~\citep{kingma2013vae}
fail to generate feasible and high quality solutions.
One approach~\citep{chao2021cgan} penalizes the generator based on constraint violation but is nontrivial to extend to more complex constraints, such as logical or general combinatorial constraints that can be readily expressed in MILP. Alternative approaches include postprocessing solutions~\citep{zhang2020milpgan}, which may lead to repetitive solutions  since the generative network is not trained with the postprocessing, or rejection sampling~\citep{gomes2006rejection} that can be computationally inefficient due to high rejection rates. Fundamentally, the training procedure does not involve verifiably feasible solutions. 
Another approach is to leverage integer linear programming (ILP) to find feasible solutions to problems with linear constraints on integer variables. 
However, such solvers are not geared towards generating diverse solutions and cannot leverage deep learning models to capture the data distribution or quickly adapt to slightly modified settings.  

In this work, we propose \genco{}. Instead of directly optimizing $\vx$, we learn a \emph{latent design space} $\cC$ so that (1) each element $\vc \in \cC$ yields a guaranteed feasible solution $\vx\in\Omega$, often through a combinatorial solver $\vg$: $\vx = \vg(\vc)$, and (2) a generative model $G_\theta$ can map noise to the latent space $\cC$. The entire pipeline is trained in an end-to-end manner via backpropagation through the solver $\vg$. Intuitively, this latent space $\cC$ represents the hidden parameters that lead to the solution. For example, $\vc$ represents the layout of a 10x10 maze, while $\vx$ represents the optimal path of the maze.

\genco{} addresses many challenges in previous work. \genco{} generates solutions efficiently due to its learned generative model $G_\theta$. Additionally, \genco{} guarantees feasibility via re-parameterization in the latent design space, which is different from conventional deep generative models like Generative Adversarial Networks (GANs) that directly generate the ultimate object of interest $\vx$, be it images or any other complex entity. \genco{} encourages diverse solutions, thanks to the group loss term $\cL$ imposed on the solution population and end-to-end training of the entire pipeline. Finally, \genco{} does not require linear objective functions and can handle nonlinear objectives. A comparison is summarized in ~\autoref{tab:summary_comparison}.  

Our main contributions involve introducing a framework, \genco{}, that can:
1) train generative models to generate solutions that are guaranteed to satisfy combinatorial properties throughout training;
2) optimize nonlinear objectives;
3) flexibly handle a variety of generative models and optimizers. 
We showcase the effectiveness of our approach empirically on various generative combinatorial optimization problems: creating game level (~\autoref{s:zelda-expts}), generating maps for path planning (~\autoref{s:warcraft}), and designing inverse photonic devices (~\autoref{s:inverse-photonic}).

\section{Related Work}

The interaction between generative models and combinatorial optimization has drawn attention as practitioners seek to generate objects with combinatorial optimization in mind. 
 
 \paragraph{Traditional constrained object generation}
 Traditional constraint optimization methods were modified to search for multiple feasible solutions for problems concerning building layout \citep{bao2013genlayout}, structural trusses \citep{hooshmand2016generativetruss}, networks \citep{peng2016networkdesign}, building interiors \citep{wu2018miqp}, and urban design \citep{hua2019ipdesign}. Additionally, an approach based on Markov chain Monte Carlo \citep{yeh2012synthesizing} samples objects that satisfy certain constraints. These approaches employ traditional sampling and optimization methods such as mathematical programming and problem-specific heuristics to obtain multiple feasible solutions. These methods often obtain multiple solutions by caching the feasible solutions found during the search process or modifying hyperparameters such as budgets or seed solutions. These methods can guarantee feasibility and optimality; however, they cannot synthesize insights from data such as historical good design examples or generate unstructured objects like images that are difficult to handle explicitly in optimization problems. 

\paragraph{Infeasibility penalization}
Further work has modified deep generative models such as GAN \citep{goodfellow2014gan} to penalize constraint violation. These methods can generate objects respecting constraint graphs \citep{para2021generative}, and blackbox constraints \citep{di2020can}. Specialized approaches consider blackbox-constrained graphs 
such as in the design of photonic crystals \citep{Christensen2020genphotocrystal}, crystal structure prediction \citep{kim2020gencrystal}, and house generation \citep{tang2023graph,nauata2021houseganpp}. Here, feasibility is encouraged through penalties and inductive bias but is not guaranteed.

\begin{figure*}
  \centering
    \includegraphics[width=\linewidth]{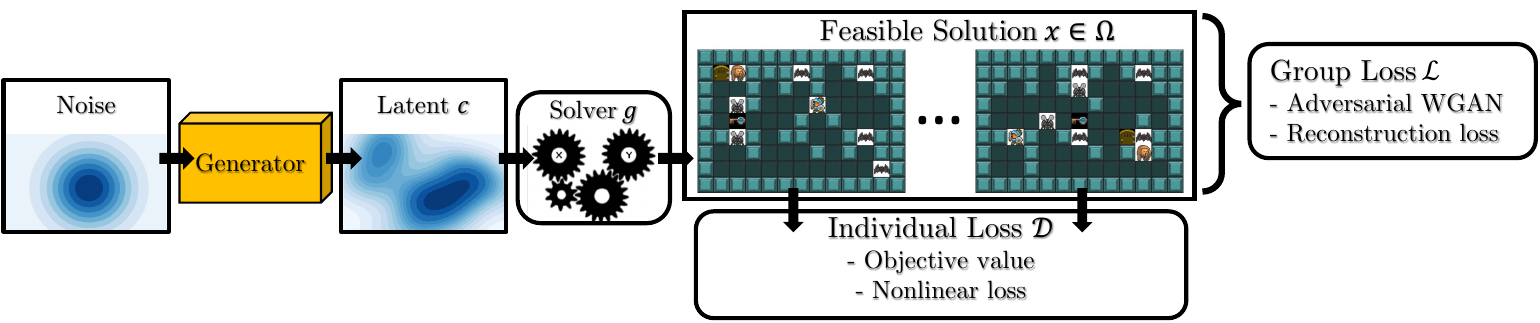} \\
  \caption{\genco{} diagram. A neural generator projects noise to a latent space which then gets used by a solver to generate provably feasible solutions. The solutions are then penalized with generative losses like Wasserstein distance (WGAN) or reconstruction (VQVAE) which can be backpropagated through the full pipeline. Additionally, \genco{} can optimize individual objectives.}
  \label{f:genco-generic}
\end{figure*}

\paragraph{Optimization-based priors}
Previous work proposed conditioning VAE \citep{kingma2013vae} with combinatorial programs \citep{misino2022vael}. Here, the latent information is extracted by the encoder and then fed through DeepProblog, a differentiable logic program~\citep{manhaeve2018deepproblog}. The result is then fed through the decoder to generate the original image based on logical relationships. At test time, the model generates objects that hopefully satisfy the logical relationship through model conditioning; however, there is no guarantee of constraint satisfaction.
Previous work has made various optimization problems differentiable, such as quadratic programs \citep{amos2017optnet}, probabilistic logic programs \citep{manhaeve2018deepproblog}, linear programs \citep{wilder2019melding, mandi2020interior, ElmachGrigas17a, liu2021risk}, Stackelberg games \citep{perrault2020game}, normal form games \citep{ling2018game}, kmeans clustering \citep{wilder2019graphs}, maximum likelihood computation \citep{niepert2021imle}, graph matching \citep{rolinek2020deep}, knapsack \citep{demirovic2019ranking, demirovic2019knapsack}, maxsat \citep{wang2019satnet}, mixed integer linear programs \citep{mandi2020spo,paulus2021comboptnet,ferber2020mipaal}, blackbox combinatorial solvers \citep{Pogancic2020diffbb, mandi2022decision, berthet2020learning}, nonlinear programs \citep{Donti_17a}, continuous constraint satisfaction \citep{donti2020dc3}, cone programs \citep{agrawal2019cvxpylayers}. General-purpose methods are presented for minimizing downstream regret by learning surrogate loss functions \citep{shah2022decision, shah2023leaving, zharmagambetov2023landscape}. Additionally, previous work has employed learnable linear solvers to solve nonlinear combinatorial problems \citep{ferber2023surco}. Finally, there is a survey on machine learning and optimization \citep{kotary2021diffoptsurvey}. These approaches focus on identifying a single solution rather than generating diverse solutions. Additionally, many of these approaches can be integrated into \genco{} if a specific optimization problem better suits the generative problem.

\paragraph{Enhacing combinatorial optimization with generative models}
Recently, generative models have been proposed to improve combinatorial optimization. \cite{zhang2022robust} use generative flow networks (gflownet) for robust scheduling problems. \cite{sun2023difusco} use graph diffusion to solve combinatorial problems on graphs. \cite{Ozair2021vaeplan} use vector quantized variational autoencoders to compress the latent space for solving planning problems in reinforcement learning. \cite{zhao2020ganrobustopt} use generative adversarial networks to generate settings for sample average approximation in robust chance-constrained optimization. Additionally, in \citep{lopez2023symmetric}, the authors generate continuous objects with tensor networks that satisfy linear constraints for optimization problems. These approaches solve fully specified optimization problems rather than generating objects with combinatorial constraints.

\def\train{\mathrm{train}}
\def\dis{\mathrm{dis}}
\def\div{\mathrm{div}}

\section{\genco{}: Method Description}
\label{s:genco-general}

\begin{table*}[htbp]
\small
\centering
\begin{tabular}{@{}r|cccccc@{}}
\toprule
\textbf{Experiment} & \textbf{Sol $\vx$} & \textbf{Feas $\Omega$} & \textbf{Group Loss $\mathcal{L}$} & \textbf{Ind. Loss $\mathcal{D}$} & \textbf{Latent $\vc$} & \textbf{Solver $\vg$} \\ \midrule
Game Design & Game Level & Playability & WGAN (comb) & -- & Soft sol. & Gurobi (ILP) \\
Path Planning & Min Path & Path & WGAN & Min path (comb) & RGB map & Gurobi (LP) \\
Photonic Device & 0/1 Grid & Manufacturing & VQVAE (comb) & Maxwell's sim. (comb) & Soft sol. & Domain spec. \\ \bottomrule
\end{tabular}
\caption{Summary of experimental settings highlighting the broad applicability of \genco{} to settings with various optimizers, generative models, and loss functions. Losses operating on combinatorial objects rather than continuous latent space are highlighted with (comb).}
\label{tab:experiment_summary}
\end{table*}

In the \genco{} framework (~\autoref{f:genco-generic}), instead of direct optimization of the solution $\vx$, we learn a distribution over a \emph{latent design space} $\cC$ so that for any $\vc\in \cC$, we get a solution $\vx = \vg(\vc) \in \cX$ that is guaranteed to be feasible. Here $\vg(\cdot)$ is a combinatorial solver that uses its input $\vc$ to specify the combinatorial optimization problems that it solves. For example, $\vg(\vc) := \argmin_{\vx\in \Omega} \vc^\top\vx$ outputs the solution to a mixed integer linear program (MILP) with coefficients $\vc$. $\vg(\cdot)$ can also output a shortest path (via Dijkstra's algorithm), given a weighted graph represented by $\vc$, etc.  

During the training, we learn a generative model $G_\theta$ so that $\vc_j = G_\theta(\epsilon_j)$ leads to solutions that minimize the loss:
\begin{equation}
    \min_{\cC := \{\vc_j\}} \cL(\{\vg(\vc_j)\}) + \gamma \sum_j \cD(\vg(\vc_j)) \quad \mathrm{s.t.}\ \vc_j = G_\theta(\epsilon_j) \label{eq:formulation2}
\end{equation}
Here $\epsilon_j$ is random noise driving generation. This can be written as the following (with $\cX := \{\vx_j\}$ representing a group of generated solutions):
\begin{equation}
    \min_\theta \cL(\cX) + \gamma\sum_j \cD(\vx_j)
    \quad\mathrm{s.t.}\ \vx_j=\vg(G_\theta(\epsilon_j)) \label{eq:formulation-ml}
\end{equation}
The generative model parameters $\theta$ can be learned via backpropagation through the combinatorial solver $\vg(\cdot)$.

\textbf{Choices of group loss $\cL$}. The group loss $\cL$ encourages the generated examples to capture the data distribution and can be instantiated using the Wasserstein distance as in Wasserstein GAN \cite{arjovsky2017wgan}, or Evidence Lower Bound (ELBO) as in VQVAE \cite{van2017vqvae}. In the WGAN case, the adversarial discriminator is trained only on examples satisfying the combinatorial constraints ($\vx$) rather than continuous solutions. Similarly, in VQVAE, the known feasible solutions are passed through the autoencoder followed by the solver $\vg$ to ensure that the autoencoder loss is only computed on feasible solutions $\vx$ rather than the continuous latent space $\vc$.

\textbf{Choices of individual loss $\cD$}. We design the individual loss $\cD$ to control the quality of individual solutions (e.g., the manufacture cost of a generated molecule $\vx_j$ should be minimized). In many settings, this individual loss is nontrivial to encode in combinatorial solvers like Gurobi as in \cite{ferber2023surco}.

\textbf{Loss on solution space $\Omega$ vs latent space $\cC$} Note that to optimize \genco{}'s objective (~\autoref{eq:formulation-ml}), we would need to run the combinatorial solver at every iteration, which involves calling MILP solvers and can be slow. On the other hand, if the group loss $\cL$ or individual loss $\cD$ can be evaluated in the latent space $\cC$, then this extra cost is not needed during training. This occurs in the map generation experiments (~\autoref{s:warcraft}), in which an observation $\vc_i$ is provided (as an RGB image) and the optimal solution $\vx_i$ can be computed accordingly (e.g., via an edge cost estimator and shortest path solver):
\begin{eqnarray}
    \min_\theta & \cL(\cC) + \gamma\sum_j \cD(\vx_j) \label{eq:formulation-ml-pen} \\\
    \ \mathrm{s.t.} & \ \vx_j=\vg(\vc_j)\quad \text{ and }\quad \vc_j = G_\theta(\epsilon_j)  \nonumber
\end{eqnarray}
This formulation is related to SurCo-zero~\cite{ferber2023surco}, in which a latent representation $\vc$ is optimized so that its inferred solution $\vx = \vg(\vc)$ by the solver $\vg$ minimizes a nonlinear loss function $\cD$. Mathematically, the training objective of SurCo is $\min_\vc \cD(\vg(\vc))$. When there are additional problem descriptions $\vy_i$ (i.e., SurCo-prior), the latent design vector $\vc_i = \vc_\theta(\vy_i)$ becomes a function of $\vy_i$ and thus the SurCo objective becomes $\min_\theta \sum_i \cD(\vg(\vc_\theta(\vy_i));\vy_i)$. The main difference here is that \genco{} considers the group loss $\cL$ (e.g., matching the data distribution) on newly generated solutions,
while SurCo focuses on the quality of solutions to known problem descriptions during training.

\iffalse
$\vx_j$ may represent a specific design and its cost may be related to a (second-level) solution to that design. For example, $\cD_2(\vx_j)$ makes sure the length of an optimal path $\vp_j = \arg\min_\vp \mathrm{MAP}(\vx_j)$ is within certain ranges, given the generated game design $\vx_j$. 

As a trade-off, this requires that we compute latent representation $\vc_i$ for the known solutions $\vx_i$ from the training set $\cX_\train$ so that a discriminator can be learned and leveraged in $\cC$. \yuandong{Did we do that??} 
\fi

%
%

%
%
%
%
%
%
%
%

%
%

In addition to ~\autoref{f:genco-generic}, we summarize our proposed method in Algorithm~\ref{alg:genco-general}. In each iteration, a set of solutions is generated using generator $G_{\theta}$ and a solver $\vg$ (i.e., forward pass). We then evaluate the loss function using \eqref{eq:formulation-ml} (or \eqref{eq:formulation-ml-pen}). The parameters of $G_{\theta}$ are updated via backpropagation. Here, the differentiation through the combinatorial solver is achieved approximately by employing methods outlined in \citep{Pogancic2020diffbb,Sahoo2022identity}.

\paragraph{Assumptions} \genco{} has several assumption on the individual loss $\mathcal{D}$, group loss $\mathcal{L}$, solver $g$, and latent space $c$. We assume the individual loss $\mathcal{D}$ is a function that maps a fixed solution to a loss value and provides gradient information useful for gradient descent. Note that if gradient information is not available, a neural network estimator may be learned to approximate the individual loss function. The group loss $\mathcal{L}$ imposes a loss on the distribution of instances generated. With deep generative models, this group loss is generally imposed by using learning-based proxies that act on individual examples. The assumption overall is that $\mathcal{L}$ can be optimized via stochastic gradient descent on the generated examples. We assume the solver $g$ is a function that maps from generated latent information $c$ to a feasible solution $x \in \Omega$. Additionally, we assume that previous work has made $g$ differentiable as a function of $c$ with various techniques proposed such as those described in related work. Finally, we assume the latent space $c$ simply represents values that can be mapped to via a neural network, and that the solver is well defined on that input. 

\begin{algorithm}[htb]
\begin{algorithmic}
    \STATE \textbf{Output}: Trained generator
    \STATE Initialize generator parameters $\gengenparams$\;
    \WHILE{not converged}
        \FOR{$j = 1\ldots K$} 
        \STATE Sample a noise $\noise_j$\;
        \STATE Sample a latent description $\vc_j = G_\theta(\noise_j)$\;
        \STATE Call a solver $\vx_j = \vg(\vc_j)$\;
        \ENDFOR
        \STATE Compute $\text{loss}$ using ~\autoref{eq:formulation-ml} with $\cX := \{\vx_j\}$ and \\ $\cC := \{\vc_j\}$.
        \STATE Backpropagate $\nabla_{\gengenparams} \text{loss}$ to update $\gengenparams$\;
     \ENDWHILE
     \caption{GenCO}\label{alg:genco-general}
\end{algorithmic}
\end{algorithm}

\section{Experiments}

We test \genco{} on three applications: Zelda game level design, Warcraft path planning map generation, and diverse inverse photonic device design. Settings are summarized in \autoref{tab:experiment_summary}, and all settings involve combinatorial optimization, making the application of deep generative models nontrivial. In these experiments, \genco{} \emph{significantly outperforms the baselines, efficiently finding diverse, realistic solutions that obey combinatorial properties}, paving the way for combining combinatorial optimization with deep generative models.

\subsection{Metrics}

We evaluate generative performance with density and coverage \cite{naeem2020reliable}. Density measures, for the average fake data point, how many real examples are nearby. Coverage measures how many real data points have a fake data point nearby. Formally, $M$ fake examples $Y_j$, are evaluated against $N$ real examples $X_i$. The metrics use the ball around $X_i$, with radius $r=\text{NND}_k(X_i)$ being the maximimum distance from $X_i$ to its $k$ nearest real neighbors, to quantify generated density and coverage of the real data distribution. They are expressed as
density = $\frac{1}{kM}\sum_{j=1}^M\sum_{i=1}^N 1_{Y_j\in B(X_i, \text{NND}_k(X_i))}$, and coverage = $\frac{1}{N}\sum_{i=1}^N 1_{\exists j \text{s.t.} Y_j \in B(X_i, \text{NND}_k(X_i))}$.

\subsection{Explicitly Constrained GAN: Game Level Design}
\label{s:zelda-expts}

\input{tables/tab_game_level}

\begin{figure*}[h!]
  \centering
  \begin{tabular}{@{}r@{}c@{}c@{}c@{}c@{}}

    \raisebox{28pt}[0pt][0pt]{\rotatebox{90}{\makebox[0pt][c]{\footnotesize GAN+MILP}}}

    \includegraphics[width=0.17\linewidth]{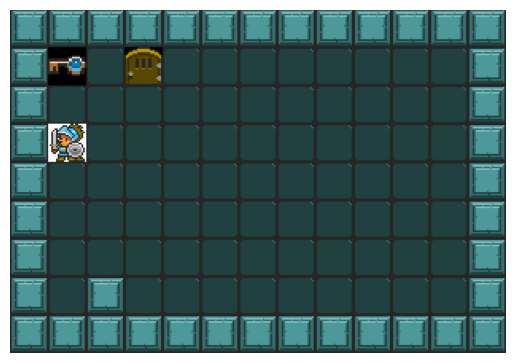} & 
    \includegraphics[width=0.17\linewidth]{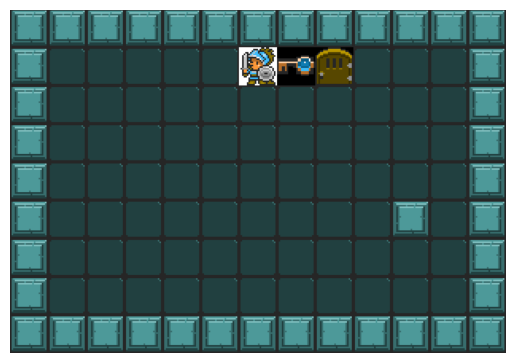} & 
    \includegraphics[width=0.17\linewidth]{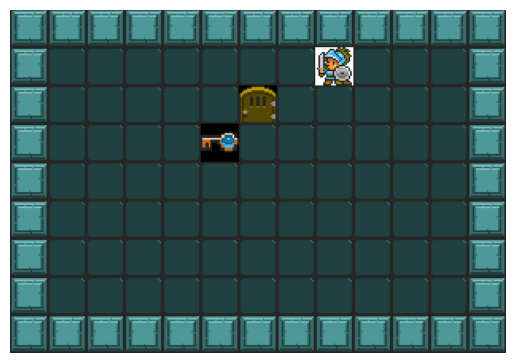} & 
    \includegraphics[width=0.17\linewidth]{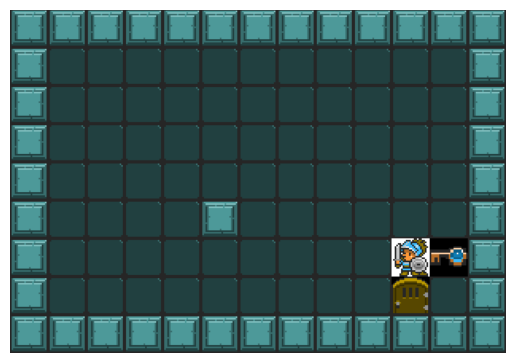} & 
    \includegraphics[width=0.17\linewidth]{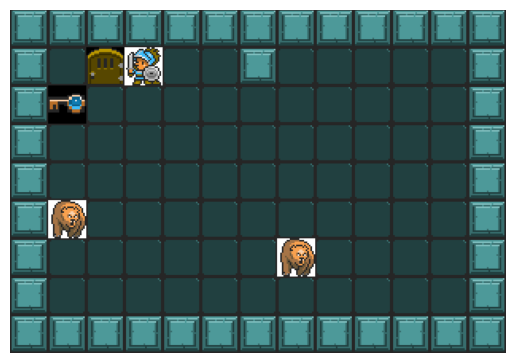}
    \\
    \raisebox{28pt}[0pt][0pt]{\rotatebox{90}{\makebox[0pt][c]{\footnotesize \genco{} fixed}}}
    \includegraphics[width=0.17\linewidth]{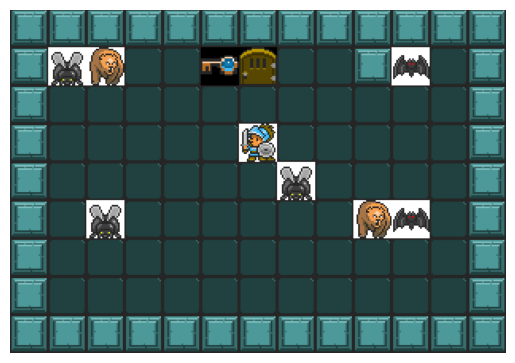} & %
    \includegraphics[width=0.17\linewidth]{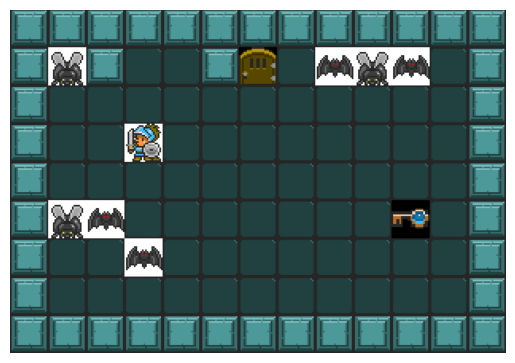} & %
    \includegraphics[width=0.17\linewidth]{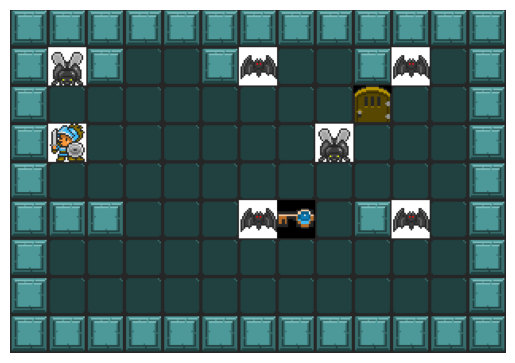} & %
    \includegraphics[width=0.17\linewidth]{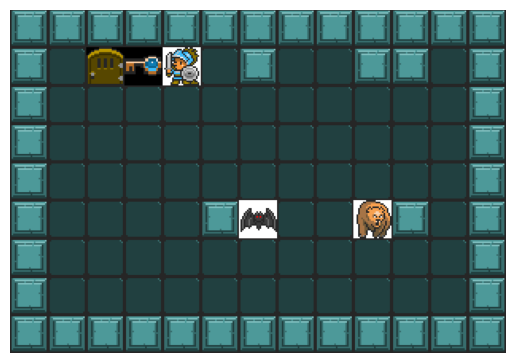} & %
    \includegraphics[width=0.17\linewidth]{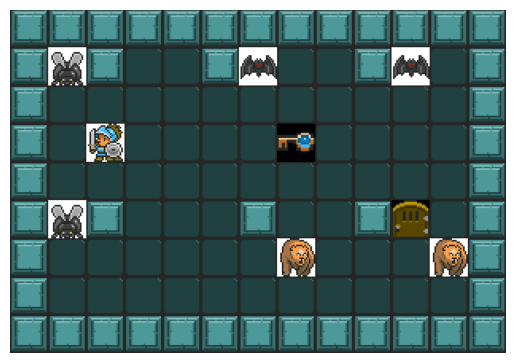}
    \\
    \raisebox{28pt}[0pt][0pt]{\rotatebox{90}{\makebox[0pt][c]{\footnotesize \genco{} upd}}}

    \includegraphics[width=0.17\linewidth]{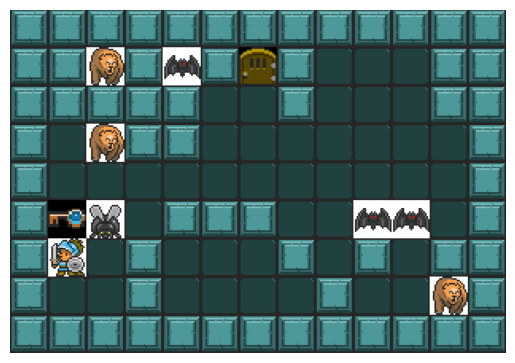} & 
    \includegraphics[width=0.17\linewidth]{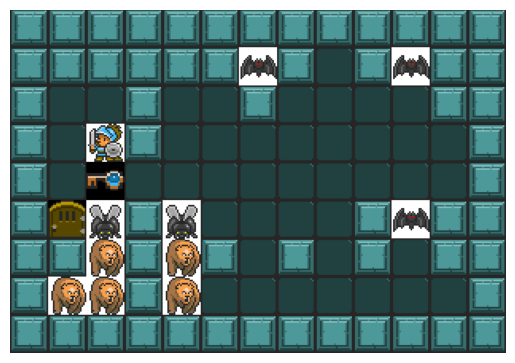} & 
    \includegraphics[width=0.17\linewidth]{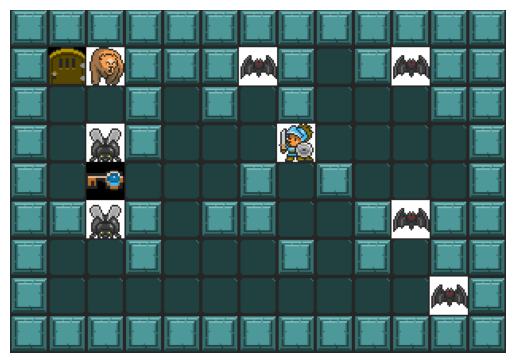} & 
    \includegraphics[width=0.17\linewidth]{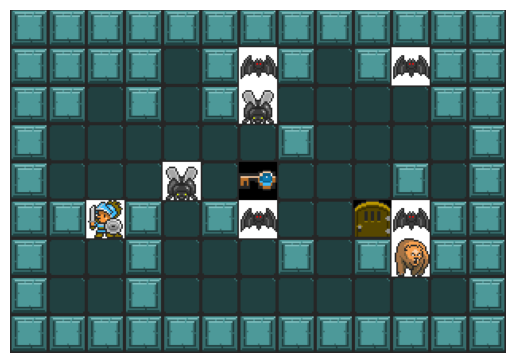} & 
    \includegraphics[width=0.17\linewidth]{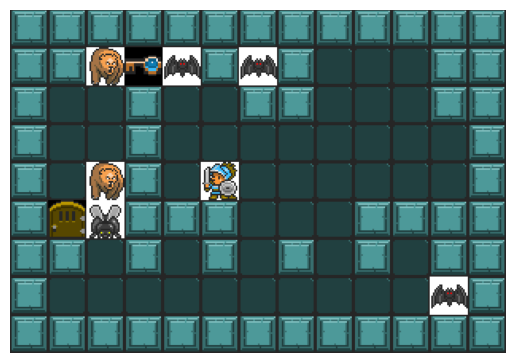}
    \\
  \end{tabular}
  \caption{Generated zelda level examples. ``\genco{} Updated'' obtains solutions that seem more realistic than the empty MILP postprocessed GAN levels and \genco{} Fixed Adversary levels.}
  \label{f:genco-zelda-viz}
\end{figure*}

We evaluate \genco{} on generating diverse and realistic Zelda game levels. Automatic level generation with playability guarantees helps designers create games with automatically generated content or iterate on development using a few examples.
In this setting, we use \genco{} to train using a \emph{constrained} WGAN formulation.
Here, we are given examples of human-crafted game levels $\cX_\train$ and are asked to generate fun new levels. The generated levels must be playable in that the player must be able to complete them by moving the character through a route that reaches the destination $(\vx\in \Omega)$. Additionally, the levels should be realistic in that they should be similar to the real game levels as measured by an adversary that is trained to distinguish between real and fake images. This adversary imposes the generative group loss $\genloss{}$, effectively approximating the Wasserstein distance \cite{arjovsky2017wgan}. We use the same 50 Zelda levels as \citep{zhang2020milpgan}, and the same neural network architecture, as we don't tune the hyperparameters for our model in particular but rather compare the different approaches, all else equal. 

\subsubsection{Baselines}

Here, we evaluate \genco{} against previous work \citep{zhang2020milpgan}(\emph{GAN + MILP fix}). This approach trains a standard Wasserstein GAN to approximate human game levels. Specifically, they train the WGAN by alternatively training a generator and a discriminator. The generator is updated to ``fool'' the discriminator (maximize the loss of the discriminator). The discriminator tries to correctly separate the generated and real game levels into their respective classes. When the practitioner wants to generate a valid level, this approach generates a soft level and fixes it using a MILP that finds the nearest feasible game level in terms of cosine distance (dot product).

We evaluate two variants of \genco{}, \emph{\genco{} - Fixed Adversary}, which trains against a fixed pretrained adversary from previous work~\citep{zhang2020milpgan}, and \emph{\genco{} - Updated Adversary}, which updates both the generator and adversary during training. Both approaches are initialized with the fully trained GAN from previous work \citep{zhang2020milpgan}. Importantly, the adversaries act on the generated solution $\vx \in \Omega$, rather than the latent soft solution $\vc$.

\subsubsection{Results}

Results are presented in \autoref{tab:genco-game-perf} and examples in \autoref{f:genco-zelda-viz}, which includes the performance of the previous approach as well as two variants of \genco{}. In these settings, we estimate performance based on sampling 1000 levels. Each level is made out of a grid, with each grid cell having one of 8 components: wall, empty, key, exit, 3 enemy types, and the player. A valid level is one that can be solved by the player in that there is a valid route starting at the player's location, collecting the key, and then reaching the exit. We evaluate the performance of the models using two types of metrics: diversity, as measured by the percentage of unique levels generated, and fidelity, as measured by the average objective quality of a fixed GAN adversary. We evaluate using adversaries from both the previous work and \genco{}. 

As shown in \autoref{tab:genco-game-perf}, we find that \genco{} with an updating adversary generates unique solutions at a much higher rate than previous approaches and also generates solutions that are of higher quality as measured by both the GAN adversary and its own adversary. The adversary quality demonstrates that the solutions are realistic in that neither the adversary from the previous work nor from \genco{} is able to distinguish the generated examples from the real examples. This is further demonstrated in \autoref{f:genco-zelda-viz} with the updated adversary generating realistic and nontrivial game levels. Furthermore, given that the levels are trained on only 50 examples, we can obtain many more game levels. Overall, \genco{} - Updated Adversary obtains the best uniqueness and generative loss. However, it obtains neither the best density nor coverage since game levels are generally sparse (the majority of tiles are empty floors), meaning that having a slightly shifted map will result in a higher distance than a completely empty map. As a result, coverage and density have counterintuitive behaviors in discrete settings while in standard image domains, samples are continuous and unconstrained. Additionally, density and coverage are relatively low. Note here that both approaches are guaranteed to give playable levels as they are postprocessed to be valid. However, \genco{} is able to generate more diverse solutions that are also of higher realism quality.

\input{tables/tab_warcraft_perf}

\paragraph{Uniqueness}
As shown in \autoref{tab:genco-game-perf}, \genco{}, with an updated adversary, obtains the highest percentage of unique solutions, generating 995 unique solutions out of 1000, significantly higher than the 520 unique solutions in previous work. This is likely due to the fact that the generator is trained with the downstream fixing explicitly in the loop. This means that while the previous work may have been able to ``hide'' from the adversary by generating slightly different continuous solutions, these continuous solutions may project to the same discrete solution. On the other hand, by integrating the fixing into the training loop, \genco{}'s generator is unable to hide in the continuous solution space and thus is heavily penalized by generating the same solution as the adversary will easily detect those to be originating from the generator. In essence, this makes the adversary's task easier as it only needs to consider distinguishing between valid discrete levels rather than continuous and unconstrained levels. This is also reflected in the adversary quality, where \genco{}'s adversary is able to distinguish between levels coming from the previous work's generator and the real levels with a much better loss.

\subsection{Map generation for path planning}
\label{s:warcraft}

In this experiment, we consider somewhat different and more challenging for \genco{} setting of generation of image maps for strategy games like Warcraft. This poses a greater challenge because the generator $G$ is not restricted and can produce any RGB image of a map (group loss $\cL$ operates directly on latent space $\cC$, i.e., $\cL(\cC)$ in eq.~\ref{eq:formulation-ml}). However, we would like to encourage $G$ to generate a map with two main criteria: 1) it should resemble real and diverse game maps, as determined by the group Loss $\cL(\cC)$; 2) \emph{the cost of the shortest path (SP) from the top-left to the bottom-right corners (source and destination) should be minimized} (group loss $\cD(\vg(\vc))$). Intuitively, the SP corresponds to (mostly) a diagonal part of the image, and a map can include various elements (terrains) like mountains, lakes, forests, and land, each with a specific cost (for example, mountains may have a cost of 3, while land has a cost of 0). To calculate this, we pass the generated image to the fixed ResNet to get the graph representation together with edge costs. This is then followed by the shortest path solver ($\vg$). The objective is to populate the map with a diverse range of objects while ensuring the shortest path remains low-cost: $\cL(\cC) + \gamma\sum_j \cD(\vg(\vc_j))$. We use the same dataset as in~\citep{Pogancic2020diffbb} with DCGAN architecture adapted from~\citep{zhang2020milpgan}. Note that the maps are unconstrained, representing the generator's latent space. However, the generated feasible solution is a valid path through the map. Additionally, the generative loss acts on the latent space $c$ rather than the feasible solution as we are given maps rather than solutions. The resnet architecture is part of the problem formulation in that it determines costs. More implementation details and experimental settings can be found in Appendix~\ref{sa:warcraft-setup}.

\subsubsection{Baselines}

We examine various baseline models, including an ``Ordinary GAN'' that does not incorporate the Shortest Path objective. As in the Zelda experiment, this approach employs a standard Wasserstein GAN architecture closely following~\citep{zhang2020milpgan}. However, in this case, we generate images directly instead of encoding game levels. More precisely, we train the WGAN by iteratively training two components: a generator and a discriminator. The generator is fine-tuned to deceive the discriminator to the greatest extent possible, aiming to maximize the discriminator's loss. The discriminator, on the other hand, endeavors to accurately distinguish between generated and authentic game levels and categorize them accordingly.

The next baseline is a ``soft'' penalty approach: ``GAN + semantic loss''. It uses a fixed NN to extract representation of the image in the form of cost vector (obtained from the fixed ResNet). This vector is then averaged and added to the final objective as a regularization. The concept is similar to the \emph{semantic loss} appeared in~\cite{di2020can}. The difference is that instead of building a circuit (via knowledge compilation) to encode the constraints, we use a fixed pre-trained ResNet as a surrogate to obtain the cost vector. While this baseline considers costs associated with objects, it does not incorporate information about the Shortest Path.

As for the \genco{}, it generalizes both of these approaches incorporating combinatorial solver into the pipeline. The detailed algorithm is given in Appendix~\ref{sa:warcraft-setup}. To backpropagate through the solver, we employ the ``identity with projection'' method from~\citep{Sahoo2022identity}. The remaining settings are similar to ``Ordinary GAN''.

\subsubsection{Results}

Quantitative results are showcased in the Table~\ref{tab:genco-warcraft-perf}. The ``Ordinary GAN'' focuses exclusively on the GAN's objective (group loss $\cL$), without considering the Shortest Path's objective (individual loss $\cD$). In contrast, \genco{} achieves higher performance with regards to the SP's objective, albeit with a slight reduction in GAN's loss. On the other hand, the ``GAN + semantic loss'' approach tries to uniformly avoid placing costly objects in any part of the image, whereas we are interested only in the shortest path. While it demonstrates a modest enhancement in the objective $\cD$, it experiences a notable decline in terms of group loss ($\cL$). This indicates a trade-off between optimizing for the Shortest Path and the GAN's objective. Here, the density and coverage are comparable between all approaches. While \genco{} doesn't obtain the highest density and coverage, it does obtain good SP loss and comparable GAN loss with Ordinary GAN, indicating a trade-off between fitting the distribution and obtaining high-quality solutions.

\begin{figure}
  \centering
  \begin{tabular}{@{}c@{}c@{}c@{}c@{}c@{}}
    \raisebox{17pt}[0pt][0pt]{\rotatebox{90}{\makebox[0pt][c]{\scriptsize Ord. GAN}}} \hspace{-1.15ex}
    \includegraphics[width=0.24\linewidth]{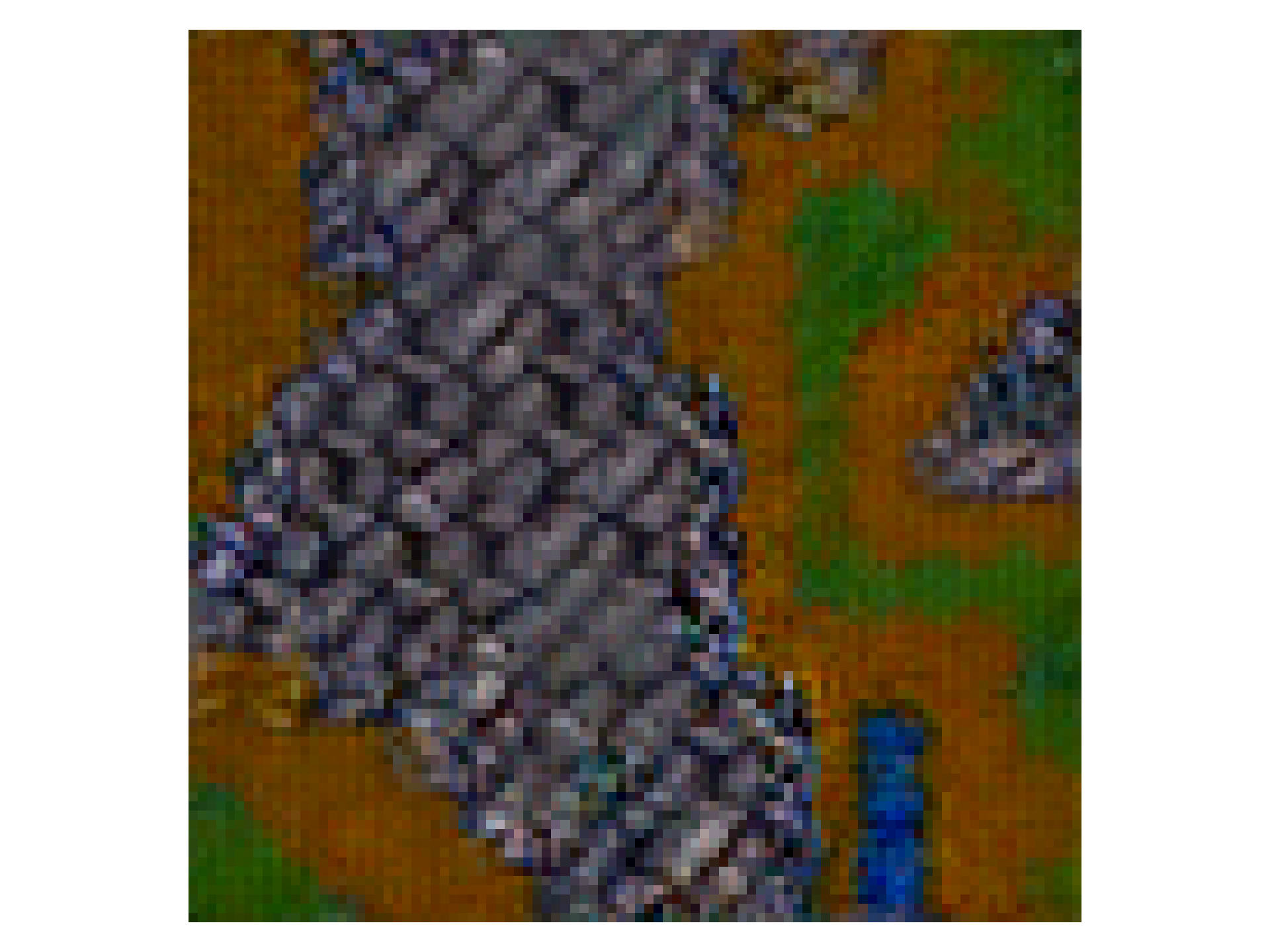} & \hspace{-3ex}
    \includegraphics[width=0.24\linewidth]{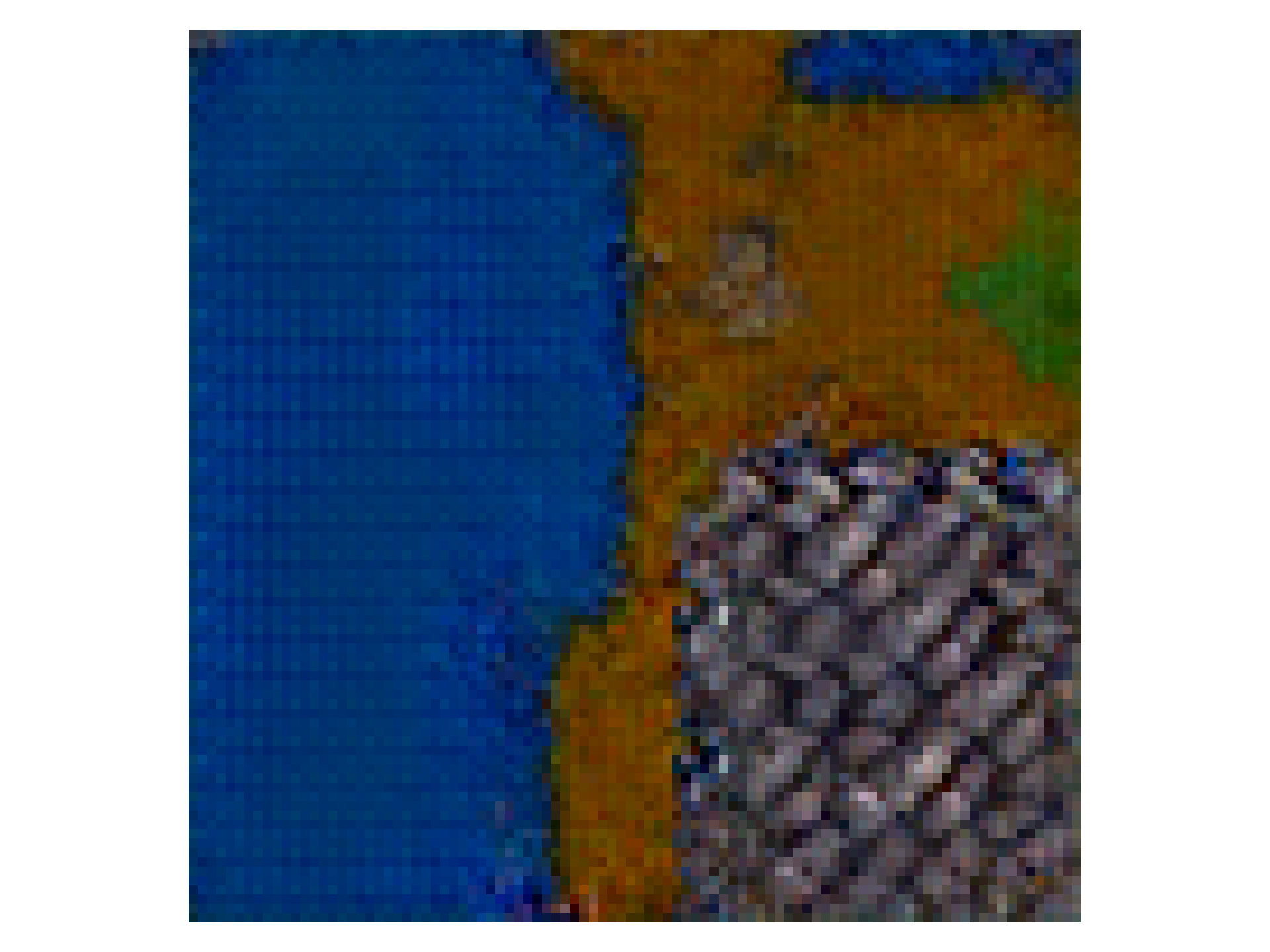} & \hspace{-3ex}
    \includegraphics[width=0.24\linewidth]{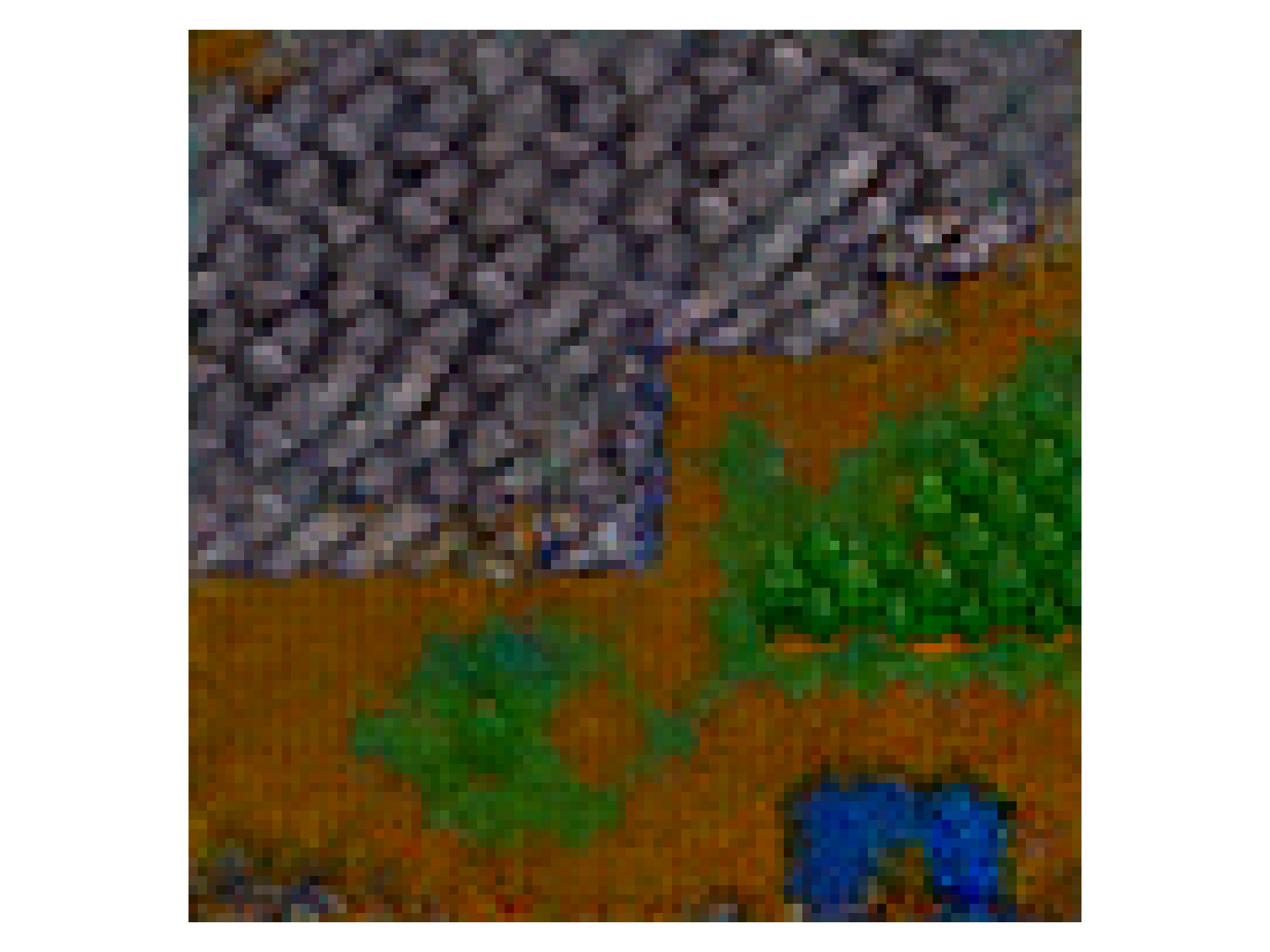} & \hspace{-3ex}
    \includegraphics[width=0.24\linewidth]{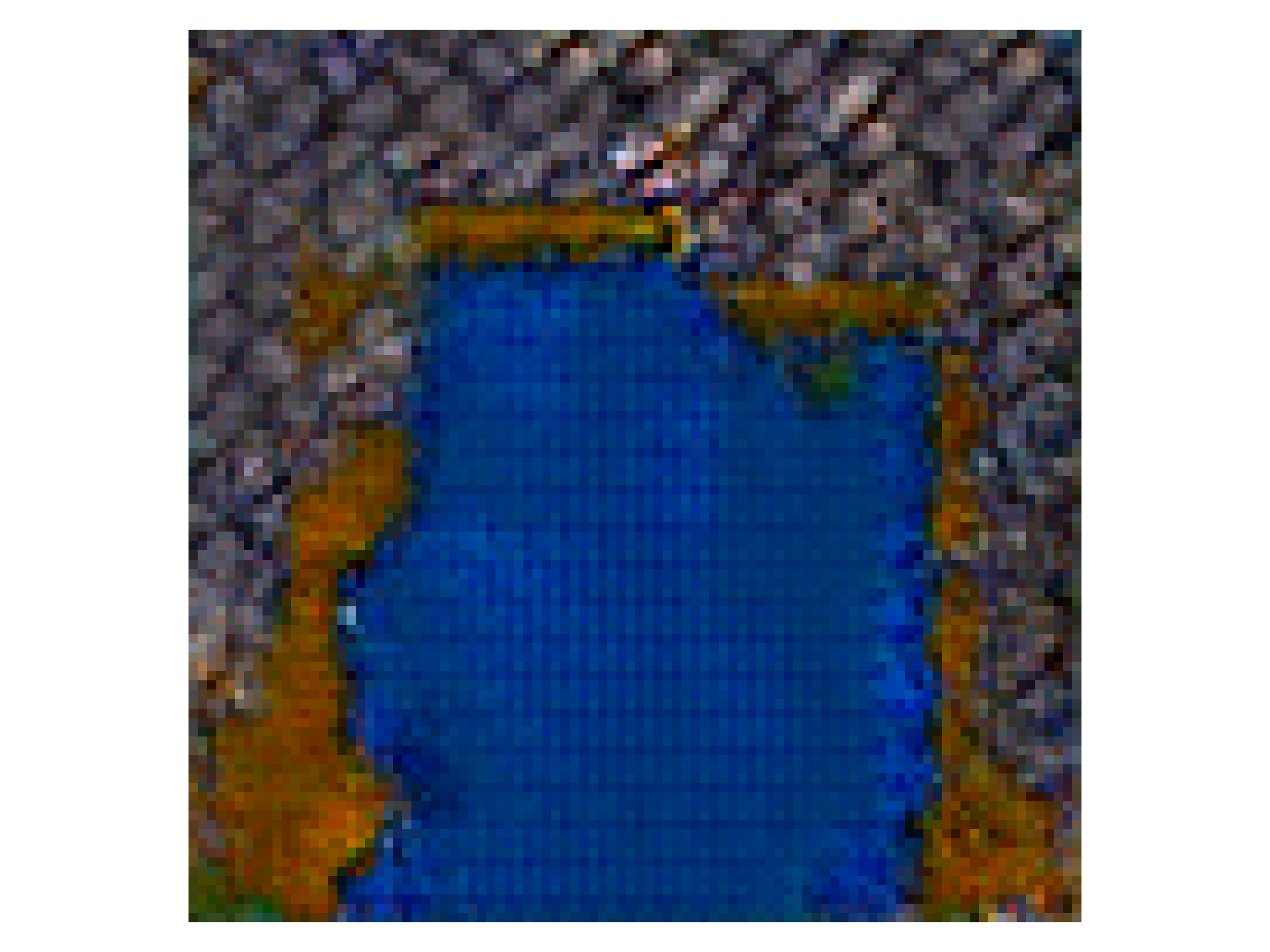} & \hspace{-3ex}
    \includegraphics[width=0.24\linewidth]{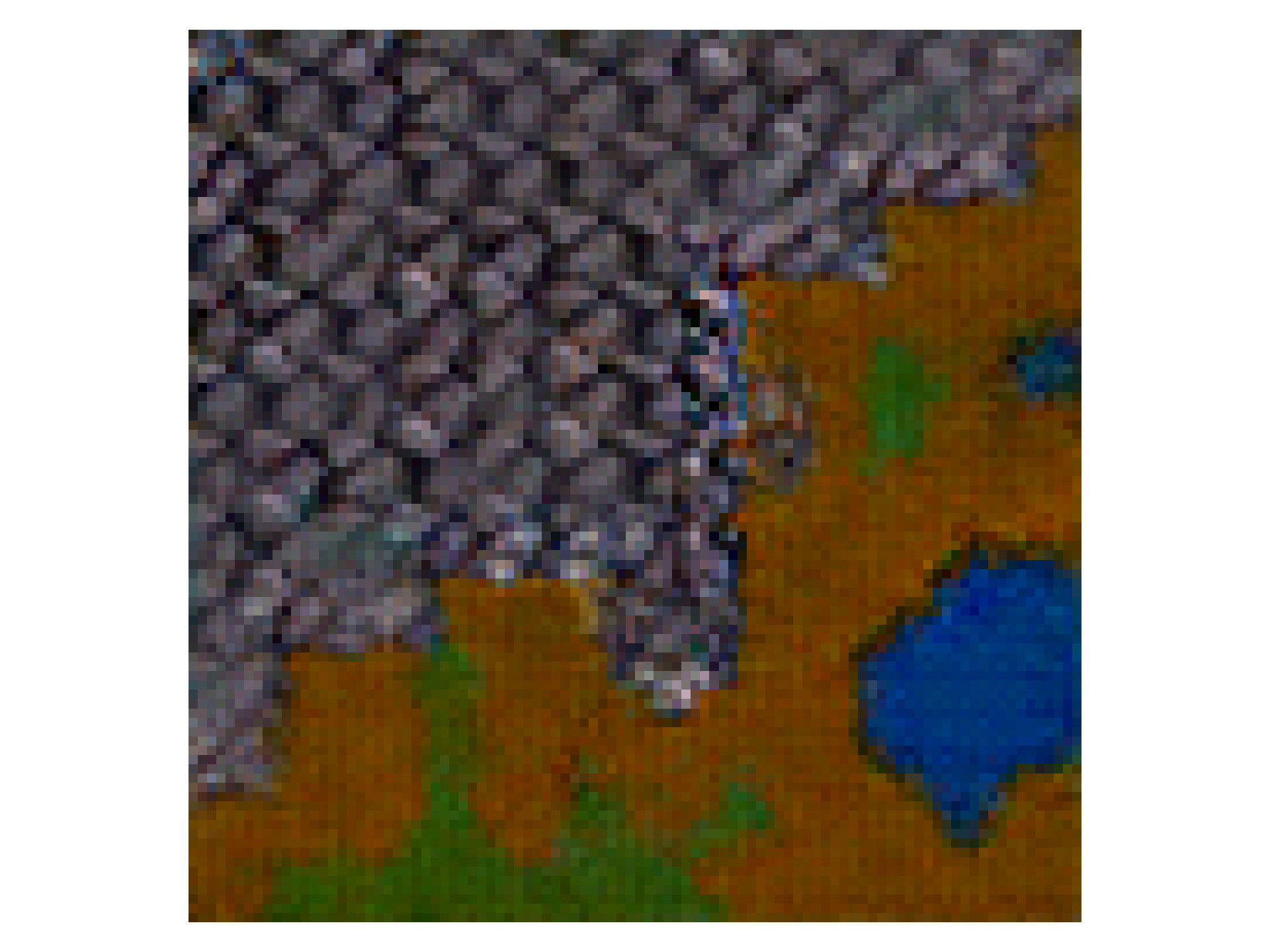}
    \\
    \raisebox{17pt}[0pt][0pt]{\rotatebox{90}{\makebox[0pt][c]{\scriptsize GAN + sem. loss}}} \hspace{-1.15ex}
    \includegraphics[width=0.24\linewidth]{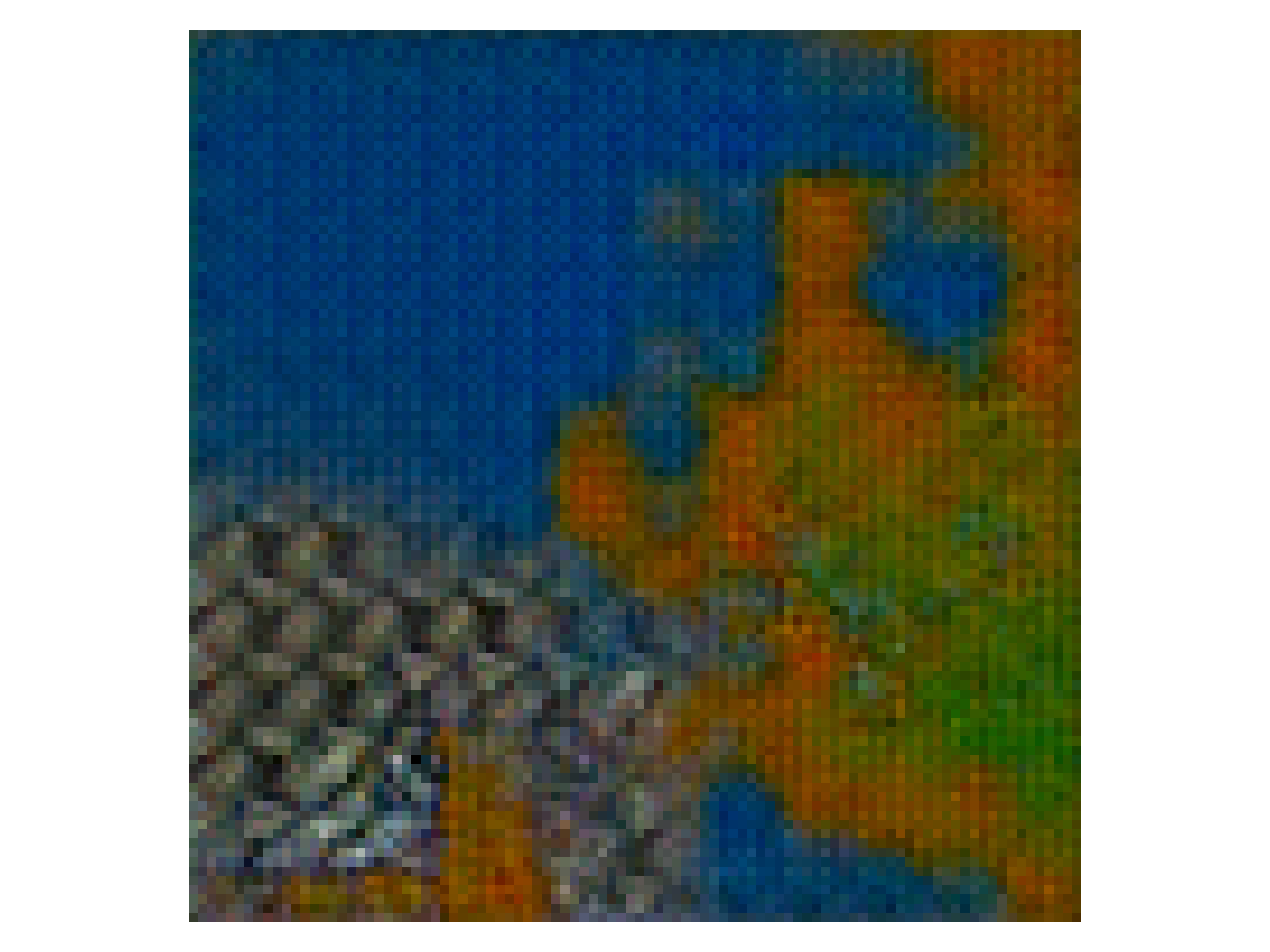} & \hspace{-3ex}
    \includegraphics[width=0.24\linewidth]{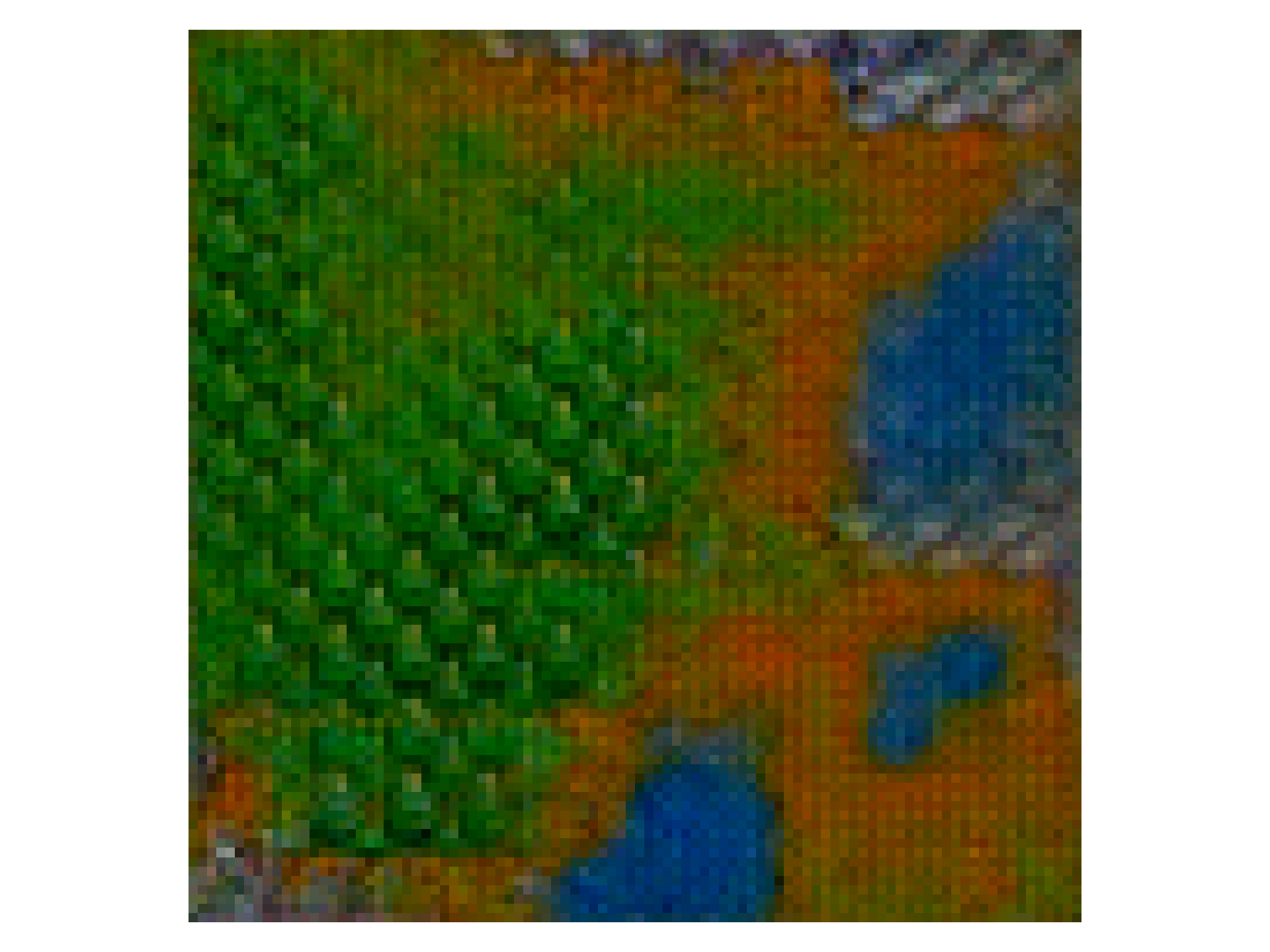} & \hspace{-3ex}
    \includegraphics[width=0.24\linewidth]{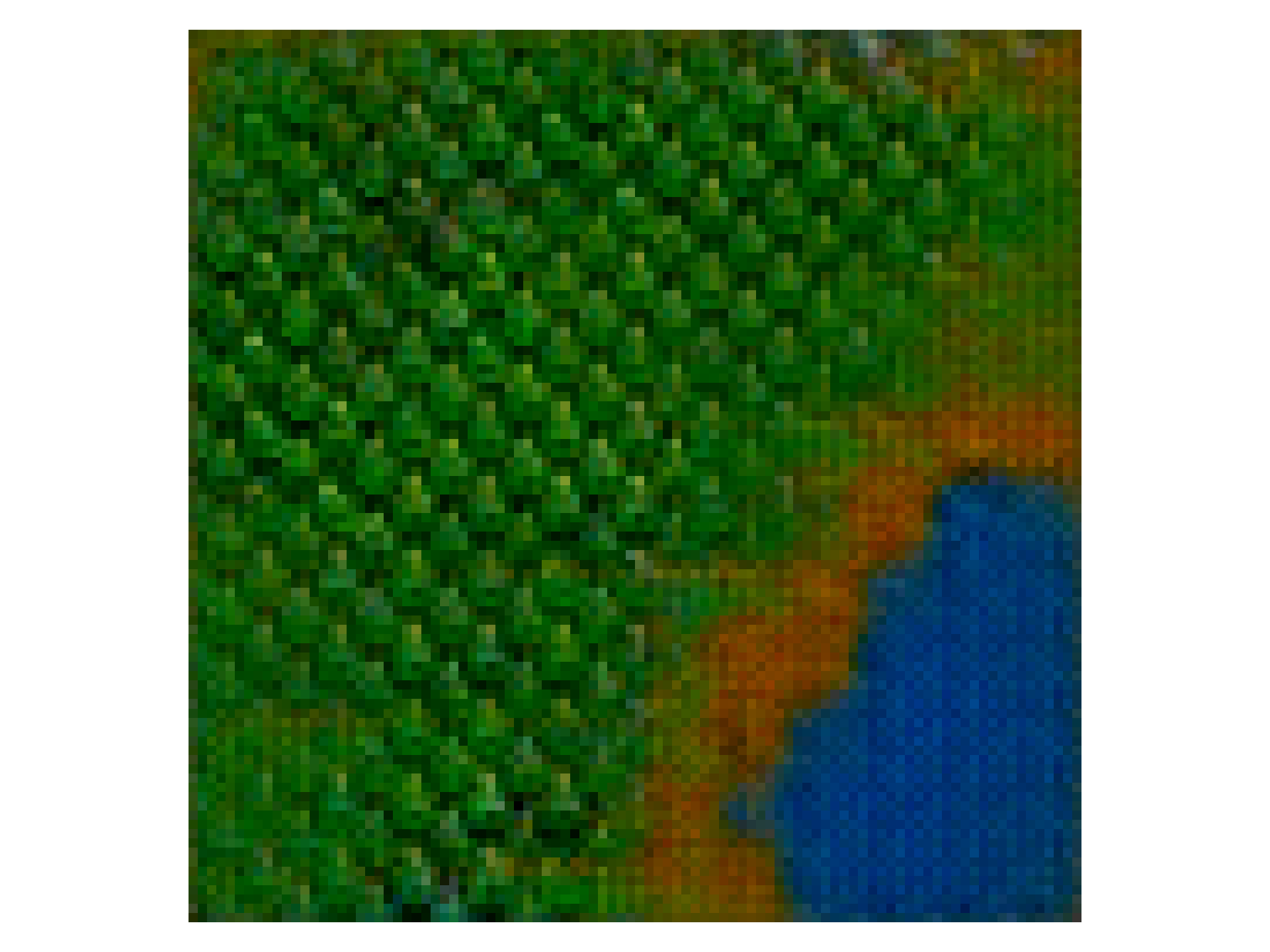} & \hspace{-3ex}
    \includegraphics[width=0.24\linewidth]{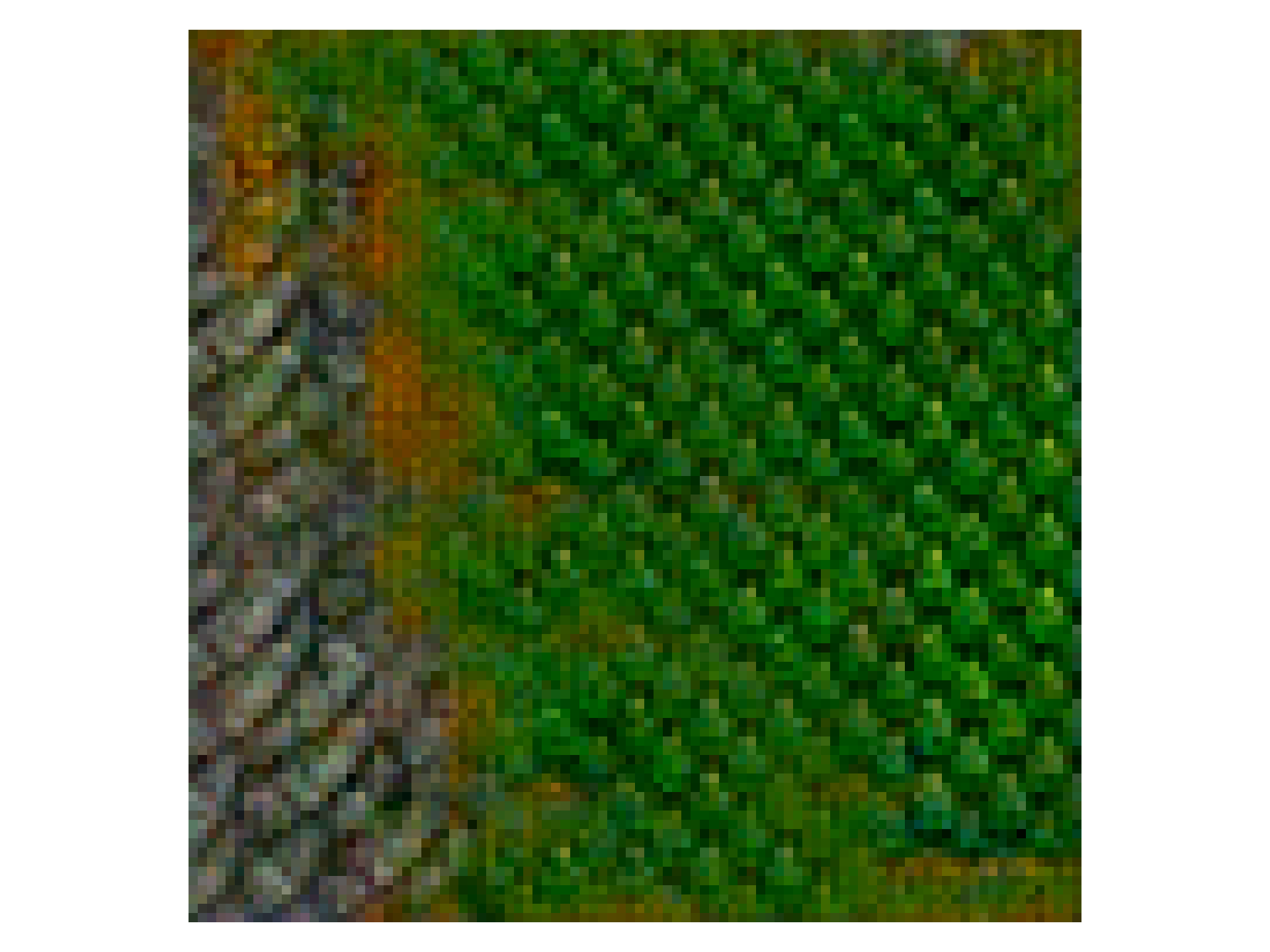} & \hspace{-3ex}
    \includegraphics[width=0.24\linewidth]{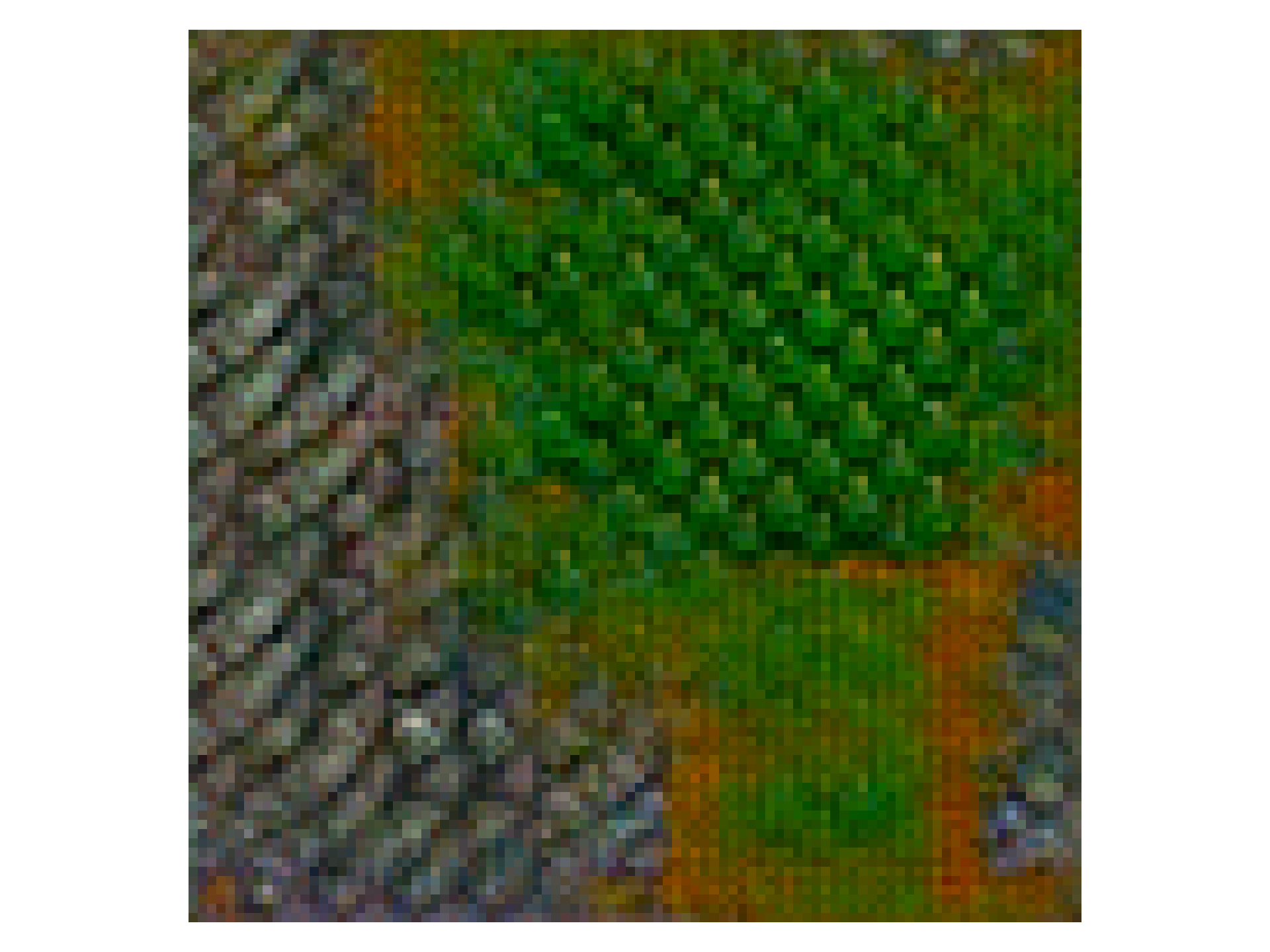}
    \\
    \raisebox{19pt}[0pt][0pt]{\rotatebox{90}{\makebox[0pt][c]{\scriptsize \genco{}}}} \hspace{-1.15ex}
    \includegraphics[width=0.24\linewidth]{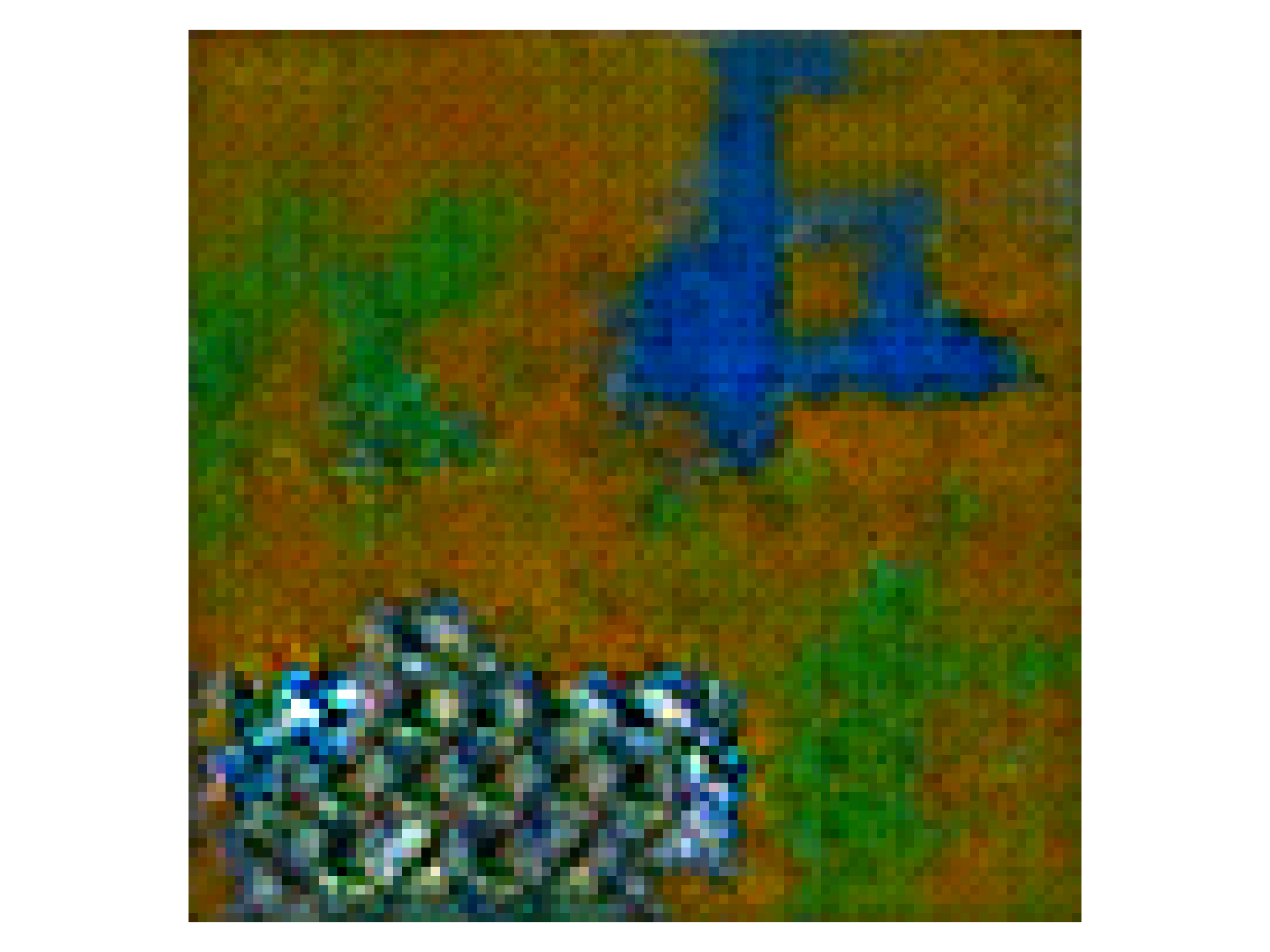} & \hspace{-3ex}
    \includegraphics[width=0.24\linewidth]{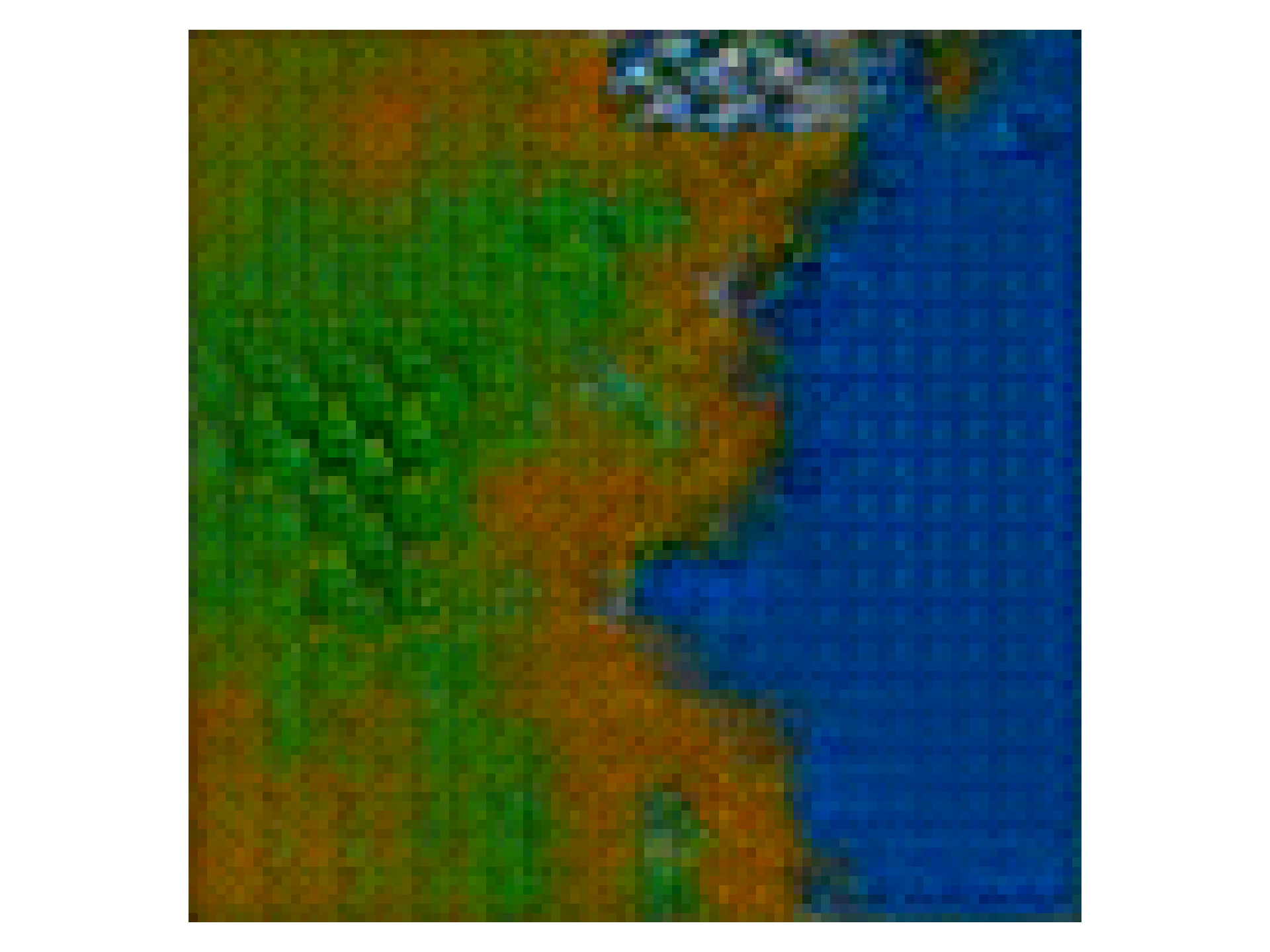} & \hspace{-3ex}
    \includegraphics[width=0.24\linewidth]{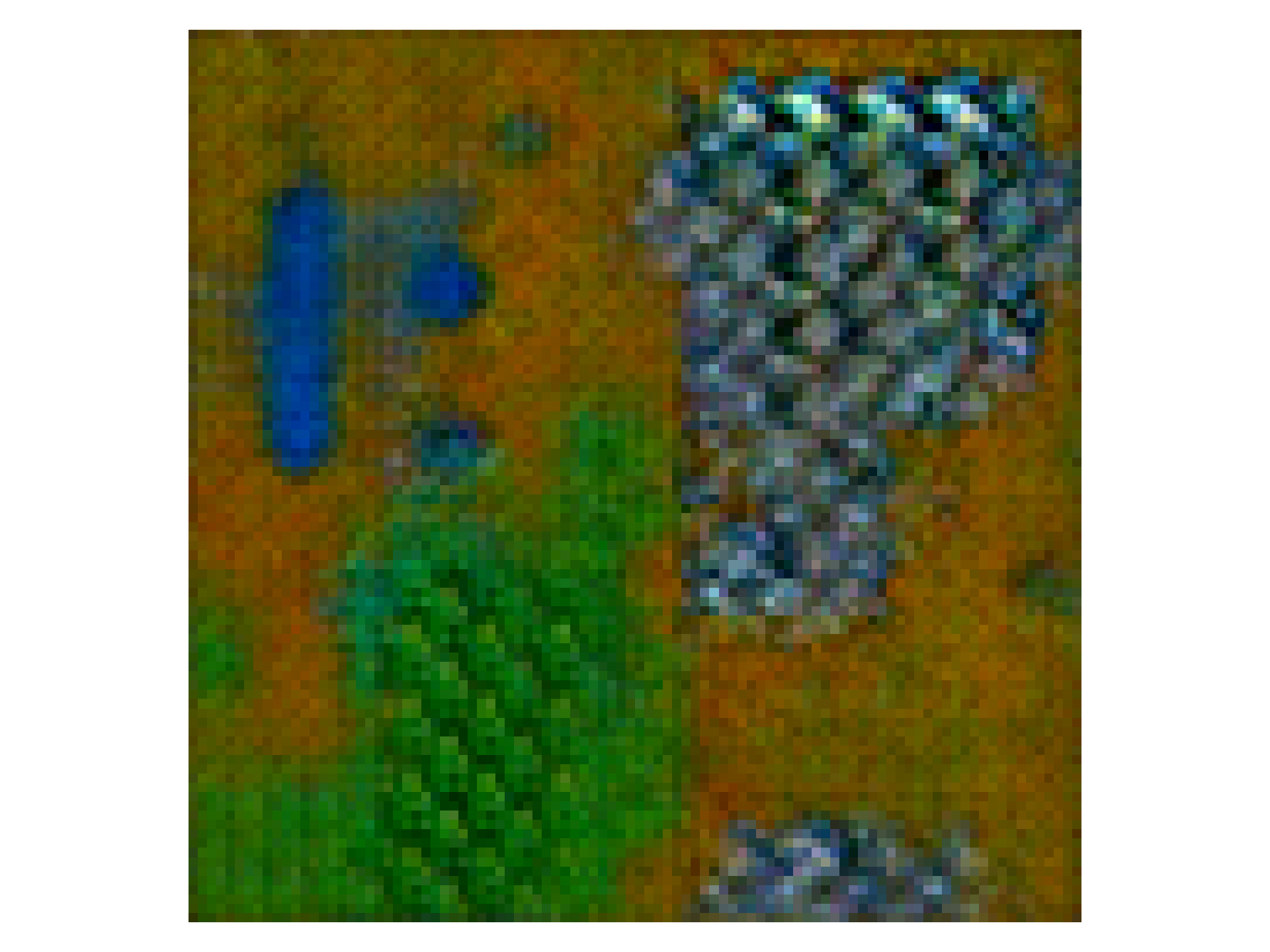} & \hspace{-3ex}
    \includegraphics[width=0.24\linewidth]{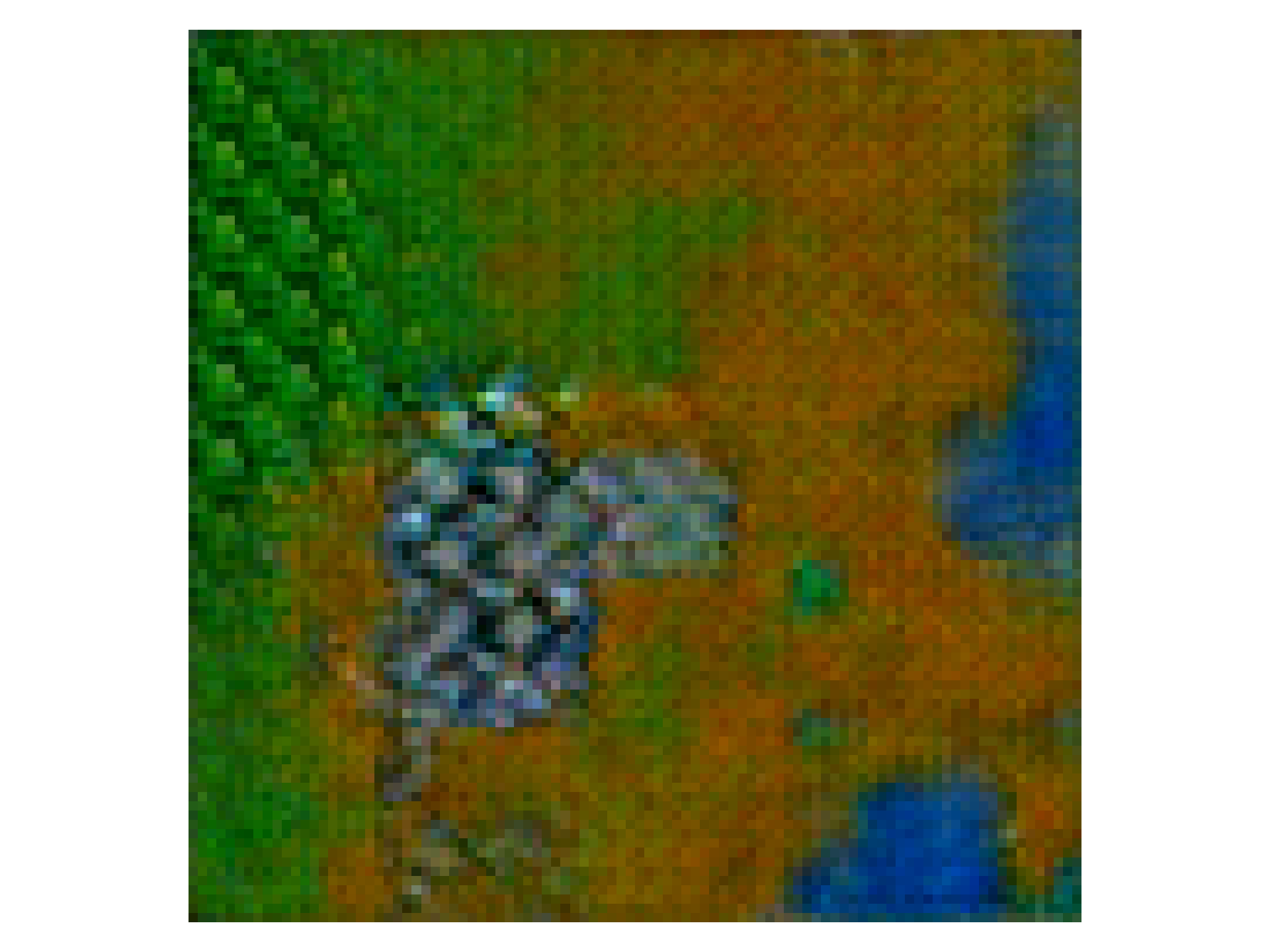} & \hspace{-3ex}
    \includegraphics[width=0.24\linewidth]{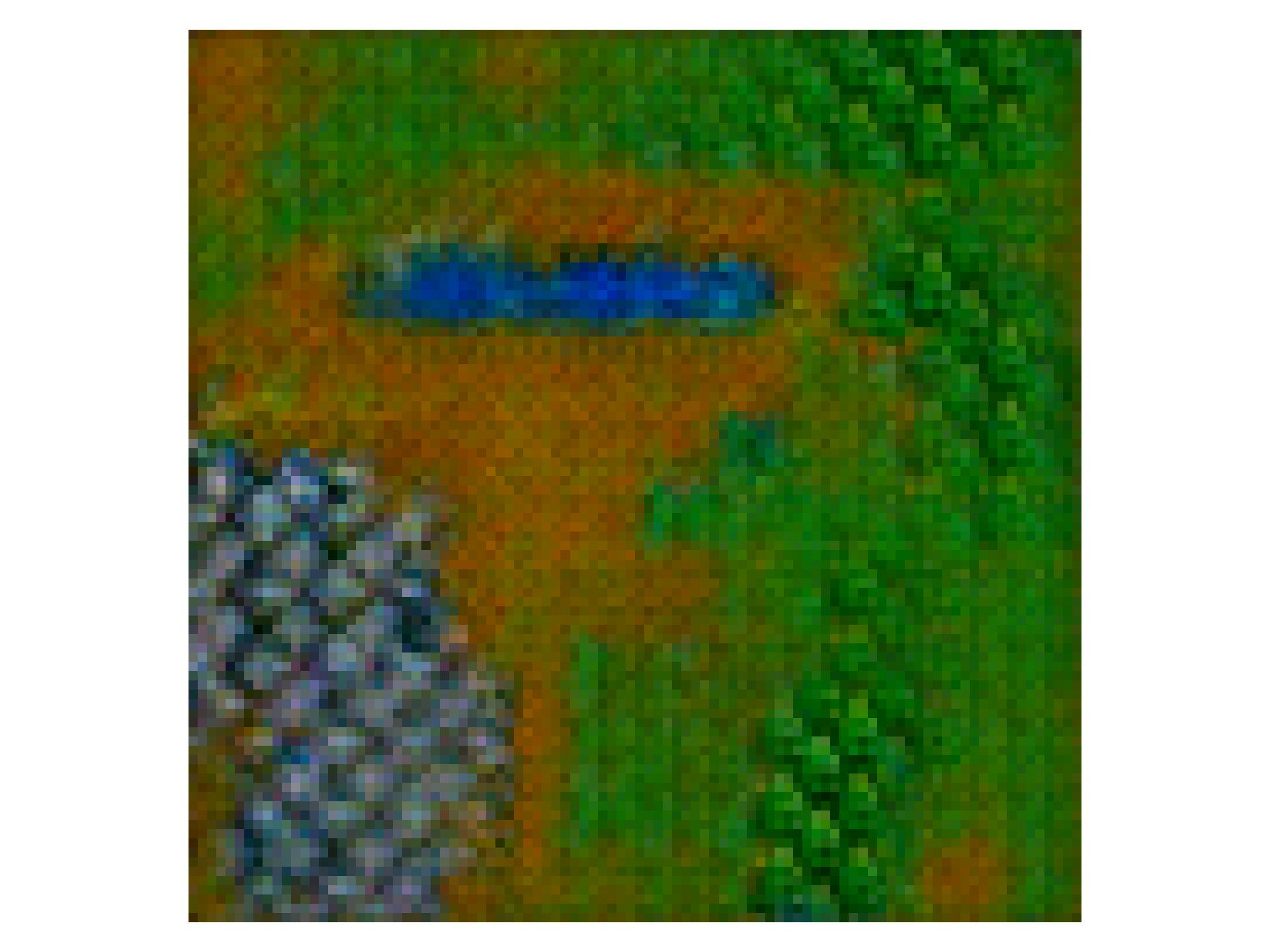}
    \\
  \end{tabular}
  \caption{A subset of generated Warcraft map images. ``Ordinary GAN'' generates ``very costly'' maps (e.g. mountains, lakes) along the Shortest Path from top-left to bottom-right; ``GAN + semantic loss'' generates less costly maps but they are \emph{less diverse}; \genco{}: SP path is cheap and the map is diverse at the same time.}
  \label{f:genco-warcraft-viz}
\end{figure}

Such quantitative results directly translate into image qualities. In fig.~\ref{f:genco-warcraft-viz}, a subset of generated Warcraft map images using various methods is displayed. The ``Ordinary GAN'' tends to generate maps with elements such as mountains and lakes, which are considered ``very costly'', particularly along the Shortest Path from the top-left to the bottom-right (which mostly goes through the diagonal). This makes sense since WGAN is trained with no information about grid costs. In contrast, the ``GAN + semantic loss'' approach produces less costly maps, albeit with reduced diversity. For instance, most part of the image is populated with the same object. On the other hand, \genco{} strikes a balance, achieving a cheap Shortest Path while maintaining a diverse range of elements on the map. The Shortest Path is efficient, and the map exhibits a rich variety of features simultaneously.

\begin{table}
\begin{tabular}{r|ccc}
\toprule
\textbf{$\gamma$} & \textbf{SP Loss ($\mathcal{D}$) $\downarrow$} & \textbf{Density $\uparrow$} & \textbf{Coverage $\uparrow$} \\
\midrule
0 & 36.45 & 0.81 & 0.98 \\
1e-4 & 27.66 & 0.93 & 0.93 \\
1e-3 & 23.99 & 0.94 & 0.93 \\
3e-3 & 23.76 & 0.75 & 0.86 \\
5e-3 & 18.02 & 0.49 & 0.61 \\
1e-2 & 0.00 & 0.00 & 0.00 \\
\bottomrule
\end{tabular}
\caption{$\gamma$ ablation, trading off individual and generative loss. High trade-off parameters improve optimization performance while low trade-off values give slightly better generation. }\label{tab:gamma_ablation}
\end{table}

Additionally, in \autoref{tab:gamma_ablation} we evaluate sensitivity to generative and shortest path loss trade-off parameter. While the shortest path loss improves with high trade-off parameter, the image quality metrics start to deteriorate rapidly. Eventually, our approach generates blank or empty images, as such images result in the shortest path with zero cost. This behavior could be attributed to the combinatorial nature of the problem.

\subsection{Constrained VAE: Photonic Design}
\label{s:inverse-photonic}

\begin{table*}[t!]
  \centering
  \begin{tabular}{r|cccc}
  \toprule
  \textbf{Approach} & \textbf{\% Unique} $\uparrow$ & \textbf{Density} $\uparrow$ & \textbf{Coverage} $\uparrow$ & \textbf{Avg Solution Loss} $\downarrow$ \\
   \midrule
  VQVAE + postprocess & 30.6\% & 0.009 & 0.006 & 1.244 \\
  \genco{} (reconstruction only) & \textbf{100\%} & 0.148 & 0.693 & 1.155 \\
  \genco{} (objective only) & 46.6\% & 0.013 & 0.036 & \textbf{0} \\
  \genco{} (reconstruction + objective) & \textbf{100\%} & \textbf{0.153} & \textbf{0.738} & \textbf{0} \\
  \bottomrule
  \end{tabular}
  \caption{\textbf{Inverse photonic design} comparison  evaluating variants of \genco{} using VQVAE against the same model architecture which postprocesses solutions. Solution loss is evaluated using a simulation of Maxwell's equations and is optimal at 0.}
  \label{tab:genco_inverse_photonics}
\end{table*}

We finally evaluate \genco{} for training a Vector Quantized Variational Autoencoder (VQVAE) to ensure that all decoded examples satisfy combinatorial manufacturing constraints $\vx\in \Omega$, are each minimizers of a photonic loss $\cD$, and are similar to given solutions via the ELBO loss $\cL$. We evaluate \genco{} against a generative + postprocess baseline \cite{zhang2020milpgan} on the inverse photonic design setting.

The inverse photonic design problem \cite{schubert2022inverse} asks how to design a device consisting of fixed and void space to route wavelengths of light from an incoming location to desired output locations at high intensity. Here the feasible region consists of satisfying manufacturing constraints that a die with a specific shape must be able to fit in every fixed and void space. Specifically, the fixed and void regions must represent the union of the die shape at different locations. The objective function here consists of a nonlinear but differentiable simulation of the light using Maxwell's equations. Previous work demonstrated an approach for finding a single optimal solution to the problem. However, we propose generating a diverse collection of high-quality solutions using a dataset of known solutions. In practice, there are various applications for generating multiple feasible photonic devices. Firstly, there are many devices that globally minimize the objective function by matching the specification perfectly. With a generator, practitioners can select from a diverse array of solutions that already optimize both specification and satisfy manufacturing constraints, quickly generating more examples if needed. Upon inspecting the solutions, they may select one that satisfies their needs, or derive high-level insights from the diversity of viable solutions to improve the design process. Lastly, optimizing the encoded objective function over the feasible region may not necessarily result in the best solution considering unquantified components like aesthetics, further manual scrutiny, and simply tacit domain expertise. Regardless, the selected design must always be optimal and satisfy the problem constraints. As such, we seek to generate diverse optimal solutions to aid the practitioner when a single solution fails to capture the full range of possibilities.

In this setting, we train a vector quantized variational autoencoder (VQVAE) \cite{van2017vqvae}. During training, the encoder encodes a known solution and then decodes an object with the same shape as the input space, using neural networks for the encoder and decoder. The continuous decoded object is then fed into the constrained optimization layer to enforce that the generated solution is feasible. This feasible solution is then used to calculate the reconstruction loss. Furthermore, in this setting we have an optimization-based objective function that consists of the simulation of Maxwell’s equations. As such, we ablate the data distribution approximation, and penalty components of \genco{}: whether or not to train using the reconstruction loss, and whether or not to penalize generated solutions based on Maxwell’s equations. We consider a baseline of training the same generative architecture without combinatorial optimization and then postprocessing generated examples during evaluation using a combinatorial solver. At test time, we sample noise and decode to the original input space. We then obtain a feasible solution using the combinatorial solver. The results in \autoref{tab:genco_inverse_photonics} include the solution uniqueness rate, the average loss evaluated using Maxwell’s equations, as well as the density and coverage.
The dataset is obtained by expensively running previous work \citep{ferber2023surco} until we obtain 100 optimal solutions. We evaluate performance on generating 1000 feasible examples.

These results in \autoref{tab:genco_inverse_photonics} demonstrate that postprocessing obtains very few unique solutions which all have high loss and furthermore don’t resemble the data distribution. This is largely due to the method not being trained with the postprocessing end-to-end. Although it closely approximates the data distribution with continuous and infeasible objects, when these continuous objects are postprocessed to be feasible, they are no longer representative of the data distribution and many continuous solutions collapse to the same discrete solution. Additionally, \genco{} obtains high density and coverage compared to the baseline postprocessed VQVAE indicating that training end-to-end helps in this setting.

\genco{} - reconstruction only gives many unique solutions that resemble the data distribution. However, the generated devices fail to perform optimally in the photonic task at hand. Disregarding the reconstruction loss and only training the decoder to generate high-quality solutions yields high-quality but redundant solutions. Combining both the generative reconstruction penalty as well as the nonlinear objective, \genco{} generates a variety of unique solutions that optimally solve the inverse photonic design problem and are representative of the data distribution.

\section{Conclusion}

In this paper, we introduce \genco{}, a framework for integrating combinatorial constraints in a variety of generative models, and show how it can be used to generate diverse combinatorial solutions to nonlinear optimization problems. Unlike existing generative models and optimization solvers, \genco{} guarantees that the generated diverse solutions satisfy combinatorial constraints, and we show empirically that it can optimize nonlinear objectives.

The underlying idea of our framework is to combine the flexibility of deep generative models with the guarantees of optimization solvers. By training the generator end-to-end with a combinatorial solver, \genco{} generates diverse and combinatorially feasible solutions, with the generative loss being computed only on feasible solutions.

Currently, many of \genco{}'s requirements are tied to those of the space of differentiable solvers which is a rapidly evolving field. Some of the key bottlenecks researchers are working to address are scalability, differentiability, and the range of optimization problems that can be differentiated. Another frontier of differentiable solver research is in learning constraints from data. Currently it is assumed that practitioners have access to a mathematical representation of the constraints that can be optimized over. However, in some domains such as molecular design, the constraints such as binding affinity are not available analytically and it is nontrivial to guarantee feasibility using mathematical programs. Learning mathematical representations of these constraints from data is an active area of research.

Additionally, as with many generative models, it is important to keep real-world deployment in mind. Even though \genco{} can guarantee feasibility with respect to explicitly encoded constraints, it cannot yet guarantee feasibility with respect to constraints that are desirable but not explicitly encoded. For instance, unless explicitly encoded or leveraging other alignment research, \genco{} cannot guarantee fairness, privacy, or positive social impact. Similarly, in image generation, it may be difficult to encode some constraints such as ensuring that pixels representing hands have exactly 5 fingers using a diffusion model in image generation. Overall, \genco{} is a step towards enforcing desirable constraints on the outputs of generative models, but it does not immediately solve all problems in ensuring the safety of generative models.

Overall, we propose \genco{}, the first generative model that generates diverse solutions, optimizing differentiable nonlinear objectives with guaranteed feasibility for explicitly combinatorial constraints. We test \genco{} on three different scenarios: inverse photonic design, game level design, and path planning, showing that \genco{} outperforms existing methods along multiple axes. We have tested \genco{} on various combinatorial optimization problems and generative settings, including GAN in Zelda game level generation, Warcraft map generation for path planning, and inverse photonic design. These settings require that \genco{} handle a variety of generative architectures (GAN / VAE), optimization problems (linear programs, quadratic programs, mixed integer programs), and solvers (SCIP, Gurobi, shortest path solvers, blackbox domain specific solvers), demonstrating the flexibility of our framework for integrating different generative paradigms and optimization methods. Our framework consistently produces diverse and high-quality solutions that satisfy combinatorial constraints, which can be flexibly encoded using general-purpose or domain specific combinatorial solvers.

\section*{Impact Statement}
This paper presents work whose goal is to advance the field of Machine Learning, deep generative models, and combinatorial optimization. There are many potential societal consequences of the fields, none which we feel must be specifically highlighted here. Additionally, while our work focuses on guaranteeing the output of generative models satisfies specified constraints, it may be difficult to characterize all types of constraints which may be impossible to specify or enumerate.

\section*{Acknowledgment}
The research at the University of Southern California was supported by the National Science Foundation (NSF) under grant number 2112533. We also thank the anonymous reviewers for helpful feedback.

%% file: tables/tab_game_level.tex
\begin{table*}[t!]
    \centering
    \begin{tabular}{r|rrrrr}
        \toprule
        Approach                        & \% Unique $\uparrow$ & Density $\uparrow$ & Coverage $\uparrow$ & GAN loss ($\genloss$) $\downarrow$ & \genco{} adversary $\downarrow$ \\
        \midrule
        GAN + MILP fix (previous)       & 0.52                 & \textbf{0.07}      & 0.94                & 0.22                              & 0.24                            \\
        \genco{} - Fixed Adversary      & 0.22                 & 0.05               & \textbf{0.98}       & -1.45                             & -0.85                           \\
        \genco{} - Updated Adversary    & \textbf{0.995}       & 0.06               & 0.82                & \textbf{-10.10}                   & \textbf{-4.49}                  \\
        \bottomrule
    \end{tabular}
    \caption{\textbf{Game level design} comparison between the previous MILP postprocessing work, \genco{} with an updating adversary, and \genco{} with a fixed adversary, with all approaches guaranteed to generate feasible levels. We measure \% Unique Density and Coverage to estimate diversity and distribution alignment. We also evaluate the adversary's prediction to gauge how well a neural network can distinguish the generated levels from real levels. }
    \label{tab:genco-game-perf}
\end{table*}

%% file: tables/tab_warcraft_perf.tex
\begin{table*}
    \centering
    \small
    \begin{tabular}{r|cccc}
    \toprule
    Approach & Density $\uparrow$ & Coverage $\uparrow$ & GAN loss ($\cL$) $\downarrow$ & SP loss ($\cD$) $\downarrow$ \\
    \midrule
    Ordinary GAN       & 0.81 & \textbf{0.98} & \textbf{0.6147} & 36.45 \\
    GAN + semantic loss~\cite{di2020can}   & \textbf{1.09} & \textbf{0.98} & 0.8994 & 35.61 \\
    \textbf{\genco{} (ours)}  & 0.94 & 0.93 & 0.6360 & \textbf{23.99} \\
    \bottomrule
    \end{tabular}
    \caption{\textbf{Map generation for Warcraft} performance comparison on 100 instances. The goal is to generate maps which are similar to the data distribution (Density / Coverage), realistic (lower GAN loss or group loss $\cL$), and have small shortest paths (SP or individual loss $\cD$).}
    \label{tab:genco-warcraft-perf}
\end{table*}

%% file: appendix.tex
\section{Detailed experimental setup for map generation (section~\ref{s:warcraft})}
\label{sa:warcraft-setup}

\subsection{Pseudocode}
\label{sa:gan-penalty}

\begin{algorithm}
\label{alg:gan-warcraft}
\begin{algorithmic}[1]
     \STATE Initialize generator parameters $\theta_{gen}$\;
     \STATE Initialize adversary parameters $\theta_{adv}$\;
     \STATE Input: distribution of the true dataset ($\vc_{true} \sim p_{\text{data}(\vc)}$), GAN's objective $\mathcal{L}(\theta_{gen},\theta_{adv})$\;
     \FOR{epoch $e$}
      \FOR{batch $b$}
        \STATE Sample a noise $\epsilon$\;
        \STATE Sample true examples from dataset $c_{true} \sim p_{\text{data}}(c)$\;
        \STATE Sample fake examples using $\vc_{fake} \sim G(\epsilon;\theta_{gen})$\;
        \STATE Transform $\vc_{fake}$ into the coefficients of the optimization problem: $\vc = f(\vc_{fake})$\;
        \STATE Solve: $x^* = \vg(\vc) = \argmin_{x\in \feas} \vc^Tx$\;
        \STATE Backpropagate $\nabla_{\theta_{gen}} \left[\mathcal{L}(\vc_{fake};\theta_{gen},\theta_{adv}) + \beta c^Tx^* \right]$ to update $\theta_{gen}$\;
        \STATE Backpropagate $\nabla_{\theta_{adv}} \left[-\mathcal{L}(\vc_{fake};\theta_{gen},\theta_{adv}) + \mathcal{L}(\vc_{real};\theta_{gen},\theta_{adv}) \right]$ to update $\theta_{adv}$\;
      \ENDFOR
     \ENDFOR
     \caption{\genco{} Penalized generator training for GANs.}
\end{algorithmic}
\end{algorithm}

Algorithm~\ref{alg:gan-warcraft} provides a detailed description of the GenCO framework in our penalty formulation and utilized in section~\ref{s:warcraft}. In this process, we sample both real and synthetic data, drawing from the true data distribution and the generator $G$ respectively (lines 7--8). Subsequently, the synthetic data undergoes a fixed mapping (e.g. ResNet in our experiments), called cost neural net (or cost NN), to obtain coefficients for the optimization problem (the edge weights for the Shortest Path). Following this, we invoke a solver that provides us with a solution and its associated objective (lines 10--11). We then proceed to update the parameters of the generator $G$ using both the GAN's objective (group loss $\cL$) and the solver's objective (individual loss $\cD$). Finally, we refine the parameters of the adversary (discriminator) in accordance with the standard GAN's objective.

\subsection{Settings}

We employ ResNet as the mapping $f(\cdot)$ from the above Algorithm~\ref{alg:gan-warcraft}, which transforms an image of a map into a $12\times12$ grid representation of a weighted directed graph: $f: \Re^{96\times96\times3} \rightarrow \Re^{12\times12}$. The first five layers of ResNet18 are pre-trained (75 epochs, Adam optimizer with lr=$5e-4$) using the dataset from~\cite{Pogancic2020diffbb}, comprising 10,000 labeled pairs of image--grid (refer to the dataset description below). Following pretraining, we feed the output into the Shortest Path solver, using the top-left point as the source and the bottom-right point as the destination. The resulting objective value from the Shortest Path corresponds to $\vg$.

\paragraph{Dataset}

The dataset used for training in the Shortest Path problem with $k=12$ comprises 10,000 randomly generated terrain maps from the Warcraft II tileset \cite{Pogancic2020diffbb} (adapted from~\cite{guyomarch17a}). These maps are represented on a $12\times12$ grid, with each vertex denoting a terrain type and its associated fixed cost. For example, a mountain terrain may have a cost of 9, while a forest is assigned a cost of 1. It's important to note that in the execution of Algorithm~\ref{alg:gan-warcraft}, we don't directly utilize the actual (ground truth) costs, but rather rely on ResNet to generate them.

\subsection{Architecture}

We employ similar DCGAN architecture taken from~\cite{zhang2020milpgan} (see fig.~3 therein). Input to generator is 128 dimensional vector sampled from Gaussian noise centered around 0 and with a std of 1. Generator consists of five (256--128--64--32--16) blocks of transposed convolutional layers, each with $3\times3$ kernel sizes and batch normalization layers in between. Discriminator follows by mirroring the same architecture in reverse fashion. The discriminator mirrors this architecture in reverse order. The entire structure is trained using the WGAN algorithm, as described in\cite{zhang2020milpgan}.

\section{GANs with combinatorial constraints}
\label{sa:genco-gan}

In the generative adversarial networks (GAN) setting, the generative objectives are measured by the quality of a worst-case adversary, which is trained to distinguish between the generator's output and the true data distribution. Here, we use the combinatorial solver to ensure that the generator's output is always feasible and that the adversary's loss is evaluated using only feasible solutions. This not only ensures that the pipeline is more aligned with the real-world deployment but also that the discriminator doesn't have to dedicate model capacity to detecting infeasibility as indicating a solution is fake and instead dedicate model capacity to distinguishing between real and fake inputs, assuming they are all valid. Furthermore, we can ensure that the objective function is optimized by penalizing the generator based on the generated solutions' objective values:
\begin{equation}\label{eq:gan-wassserstein}
    \genloss{}(G(\noise; \theta_{gen})) = \E_{\noise{}} \left[ \log(1 - f_{\theta_{adv}}(G(\noise; \theta_{gen}))) \right]
\end{equation}
where $f_{\theta_{adv}}$ is an adversary (a.k.a. discriminator) and putting this in the context of \eqref{eq:formulation} leads to:
\begin{equation}\label{eq:gan-genco}
    \begin{aligned}
    \min_{\theta_{gen}} \quad & \genloss{}(S(\unconstrainedgen(\noise; \theta_{gen}))) = \E_{\noise{}} \left[ \log(1 - f_{\theta_{adv}}(S(\unconstrainedgen{}(\noise; \theta_{gen}))) ) \right]
    \end{aligned}
\end{equation}
where $\unconstrainedgen{}$ is unconstrained generator and $\solver$ is a surrogate combinatorial solver as described above. Here, we also have adversary's learnable parameters $\theta_{adv}$. However, that part does not depend on combinatorial solver and can be trained as in usual GAN's. The algorithm is presented in pseudocode~\ref{alg:genco-gan}.

\begin{algorithm}[H]\label{alg:genco-gan}
\begin{algorithmic}
     \STATE Initialize generator parameters $\theta_{gen}$\;
     \STATE Initialize adversary parameters $\theta_{adv}$\;
     \FOR{epoch $e$}
      \FOR{batch $b$}
        \STATE Sample problem $\noise{}$\;
        \STATE Sample true examples from dataset $x_{true} \sim p_{\text{data}}(x)$\;
        \STATE Sample linear coefficients $c \sim G(\noise{};\theta_{gen})$\;
        \STATE Solve $x^* = \argmax_{x\in \feas} c^Tx$\;
        \STATE Backpropagate $\nabla_{\theta_{gen}} \genloss{}(x^*;\theta_{gen})$ to update generator (\eqref{eq:gan-genco})\;
        \STATE Backpropagate $\nabla_{\theta_{adv}} \left[ \log(f_{\theta_{adv}}(x_{true})) - \genloss{}(x^*;\theta_{gen}) \right]$ to update adversary\;
      \ENDFOR
     \ENDFOR
     \caption{\genco{} in the constrained generator setting}
\end{algorithmic}
\end{algorithm}

\section{\genco{} -- VQVAE}
\label{sa:vqvae}

The formulation below spells out the VQVAE training procedure. Here, we simply train VQVAE on a dataset of known objective coefficients, which solves the problem at hand. A variant of this also puts the decision-focused loss on the generated objective coefficients, running optimizer $g$ on the objective coefficients to get a solution and then computing the objective value of the solution.

\begin{equation}\label{eq:elbo}
    \mathcal{L}(\cX) = \text{ELBO}(\vx, \theta, E) = \mathbb{E}_{q_\theta(z|\vx)}[\log p_\theta(\vx|z)] - \beta \cdot D_{\text{KL}}(q_\theta(z|\vx)||p(z)) + \gamma \cdot \|sg(\mathbf{e}_k) - \mathbf{z}_{e,\theta}\|_2^2
\end{equation}
Here $z$ is an embedding vector, $c$ is the objective coefficients, $\log p_\theta(\vx|z)$ is a loss calculated via the mean squared error between the decoder output and the original input objective coefficients, $q_\theta(z|\vx)$ is the encoder, $p(z)$ is the prior, and $sg(\cdot)$ is the stop gradient operator, $E$ is a discrete codebook that is used to quantize the embedding.

\begin{equation}\label{eq:surco_loss}
    \mathcal{D} = \E_{c\sim p_\theta(c\mid z)}\left[f_\text{obj}(g(c; y))\right]
\end{equation}

 The algorithm below maximizes a combination of the losses in Equation~\eqref{eq:elbo} and Equation~\eqref{eq:surco_loss}.

\begin{algorithm}
\caption{Constrained Generator Training for VQVAE}
\begin{algorithmic}[1]

\STATE \textbf{Given:} Training data distribution \(D\) over problem info \(y\) and known high-quality solutions \(x\), regularization weight \(\beta\), linear surrogate solver $g_\text{solver}$, nonlinear objective $f_\text{objective}$.\\
\STATE \textbf{Output:} Trained encoder \(f_{\text{enc}}\), decoder \(f_{\text{dec}}\), and codebook \(E\)\\
\rule{\linewidth}{0.4pt} %
\STATE Initialize the parameters of the encoder \(f_{\text{enc}}\), decoder \(f_{\text{dec}}\), and the codebook \(E = \{e_1, e_2, \ldots, e_K\}\) with \(K\) embedding vectors
\FOR{\(t = 1\) {\bfseries to} \(T\)}
    \STATE Sample \(y, x\) from the distribution \(D\)
    \STATE Compute the encoder output \(z_e = f_{\text{enc}}(y, x)\)
    \STATE Find the nearest embedding vector \(z_q = \arg\min_{e \in E} \|z_e - e\|_2^2\)
    \STATE Compute the quantization loss \(L_{\text{quant}} = \|z_e - z_q\|_2^2\)
    \STATE Decode the embedding \(\tilde{c} = f_{\text{dec}}(y, z_q)\)
    \STATE Solve \(\tilde{x} = \argmax_{x\in \feas} c^Tx\)
    \STATE Compute the reconstruction loss \(L_{\text{recon}} = \|x - \tilde{x}\|_2^2\)
    \STATE Compute the optimization loss \(L_{\text{opt}} = f_{\text{objective}}(\tilde{x})\)
    \STATE Compute the total loss: \(L_{\text{total}} = L_{\text{recon}} + \beta_1 L_{\text{quant}} + \beta_2 L_{\text{opt}}\)
    \STATE Update the parameters of the encoder, decoder, and codebook to minimize \(L_{\text{total}}\)
\ENDFOR

\end{algorithmic}
\end{algorithm}